\documentclass{article}

\usepackage[final]{neurips_2024}
\usepackage{microtype}
\usepackage{graphicx}
\usepackage{booktabs} 
\usepackage{subcaption}
\usepackage{times}
\usepackage{latexsym}
\usepackage{bm}
\usepackage{threeparttable} 
\usepackage{rotating}
\usepackage{multirow}
\usepackage{multirow, makecell}
\usepackage{arydshln}
\usepackage[T1]{fontenc}
\usepackage[utf8]{inputenc}
\usepackage{microtype}
\usepackage{graphicx}

\usepackage{inconsolata}
\usepackage{subcaption}
\usepackage{setspace}
\usepackage{enumerate}
\usepackage{enumitem}
\usepackage{lipsum}
\usepackage{algorithm}   
\usepackage[noend]{algpseudocode}

\usepackage{amsmath,amsfonts,bm}

\def\eqref#1{equation~\ref{#1}}

\def\1{\bm{1}}

\def\va{{\bm{a}}}

\def\vw{{\bm{w}}}

\def\mA{{\bm{A}}}
\def\mB{{\bm{B}}}
\def\mC{{\bm{C}}}
\def\mD{{\bm{D}}}

\def\mG{{\bm{G}}}

\def\mK{{\bm{K}}}

\def\mQ{{\bm{Q}}}

\def\mS{{\bm{S}}}

\def\mV{{\bm{V}}}
\def\mW{{\bm{W}}}
\def\mX{{\bm{X}}}

\DeclareMathAlphabet{\mathsfit}{\encodingdefault}{\sfdefault}{m}{sl}
\SetMathAlphabet{\mathsfit}{bold}{\encodingdefault}{\sfdefault}{bx}{n}

\def\gC{{\mathcal{C}}}

\def\gL{{\mathcal{L}}}

\def\gO{{\mathcal{O}}}

\def\gQ{{\mathcal{Q}}}

\newcommand{\softmax}{\mathrm{softmax}}

\usepackage[accsupp]{axessibility}
\usepackage{anyfontsize}
\usepackage{multicol}

\def\aespa{{\textit{aespa}}}

\newcommand{\ie}{\textit{i.e.,}}
\newcommand{\zfold}{\textsc{Z-Fold}}
\newcommand{\brecq}{\textsc{Brecq}}
\newcommand{\eg}{\textit{e.g.,}}

\newcommand{\vect}{\operatornamewithlimits{vec}}
\newcommand{\SA}{\mathrm{SA}}
\newcommand{\trace}{\operatornamewithlimits{tr}}
\newcommand{\diag}{\operatornamewithlimits{diag}}

\makeatletter
\newcommand*\mysize{%
  \@setfontsize\mysize{10.0}{9.0}%
}
\makeatother

\usepackage{hyperref}

\usepackage{amsmath}
\usepackage{amssymb}
\usepackage{mathtools}
\usepackage{amsthm}

\usepackage[capitalize]{cleveref}
\crefname{equation}{}{}
\crefname{figure}{Figure}{Figures}
\Crefname{section}{Section}{Sections}
\Crefname{table}{Table}{Tables}

\theoremstyle{plain}

\theoremstyle{definition}

\theoremstyle{remark}

\usepackage{balance}
\usepackage[textsize=tiny]{todonotes}
\usepackage{graphicx}
\usepackage{wrapfig,lipsum}

\title{%
  \begin{tabular}{@{}c@{}l} 
    \raisebox{-0.4\totalheight}{\includegraphics[width=0.5cm]{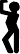}} &
    \begin{tabular}{@{}c@{}}~~Towards Next-Level Post-Training Quantization \\ of Hyper-Scale Transformers
    \end{tabular}
  \end{tabular}
}

\author{%
  Junhan Kim\thanks{Equal Contribution, $^\dagger$Corresponding Author},~~~~Chungman Lee\footnotemark[1],~~~~Eulrang Cho,~~~~Kyungphil Park,\\ \textbf{Ho-young Kim,~~~~~Joonyoung Kim,~~~~~Yongkweon Jeon}$^{\dagger}$\\
  Samsung Research\\
  \texttt{\{jun\_one.kim, chungman.lee, dragwon.jeon\}@samsung.com}
}

\begin{document}

\maketitle

\begin{abstract}
With the increasing complexity of generative AI models, post-training quantization (PTQ) has emerged as a promising solution for deploying hyper-scale models on edge devices such as mobile and TVs.
Existing PTQ schemes, however, consume considerable time and resources, which could be a bottleneck in real situations where frequent model updates and multiple hyperparameter tunings are required.
As a cost-effective alternative, learning-free PTQ schemes have been proposed. 
However, the performance is somewhat limited because they cannot consider the inter-layer dependency within the attention module, which is a significant feature of Transformers.
In this paper, we thus propose a novel PTQ algorithm that balances accuracy and efficiency.
The key idea of the proposed algorithm called \aespa \ is to perform quantization layer-wise for efficiency while targeting attention-wise reconstruction to consider the cross-layer dependency.
Through extensive experiments on various language models and complexity analysis, we demonstrate that \aespa \ is accurate and efficient in quantizing Transformer models. The code will be available at \url{https: //github.com/SamsungLabs/aespa}.
\end{abstract}

\section{Introduction} \label{sec:introduction}

Model size has been gradually growing, resulting in deep generative models such as diffusion~\cite{Rombach2022high} and large-scale language models (LLMs)~\cite{touvron2023llama,zhang2022opt}  becoming more mainstream; the trend of AI is transitioning from discriminative models to generative models with numerous parameters in trillions.
With the explosive growth in model complexity (parameters), the performance of AI models has been advancing and is now approaching or even exceeding human intelligence levels.
However, this growth in scale has resulted in a corresponding increase in computational costs, which necessitates the efficient processing and compression of AI models.
Interestingly, one attempts to expand the complexity of AI models to scale up performance, whereas the other aims to compress models to reduce cost.

Quantization is a promising solution and indispensable procedure facilitating the efficient deployment of AI models on devices that mainly support fixed-point arithmetic. 
By reducing the precision of weights, the memory bandwidth requirements can be relieved, and the embarrassing parallelism of quantized models can be SIMDified using highly efficient vector processing units such as NPU. 
To minimize the inevitable performance degradation caused by quantization, we can choose one of two approaches: quantization-aware training (QAT)~\cite{esser2019learned,jung2019learning} and post-training quantization (PTQ)~\cite{nagel2020up,li2021brecq}.
Considering the model complexity and required resources (\eg \, training costs and available datasets), QAT is not practical for compressing models with billions of parameters. 
Consequently, recent quantization studies on hyper-scale Transformer~\cite{vaswani2017attention} models have focused more on PTQ. 

Although existing PTQ schemes have successfully quantized relatively small-scale models (\eg \, ResNet)~\cite{nagel2020up,hubara2021accurate,li2021brecq,obq,jeon2022mr}, they have difficulty handling large-scale models because of their time and space complexity.
As a cost-effective alternative, learning-free algorithms have been proposed recently~\cite{frantar2023optq, jeon2023frustratingly, lin2023awq}, but their performance is somewhat limited because they do not consider the inter-layer dependency and are reliant on the nearest rounding.
There is an accuracy-efficiency trade-off; thus, we aim to bridge the gap toward next-level quantization of hyper-scale Transformer models.

In this paper, we propose a novel PTQ algorithm, called \aespa,\footnote{\aespa:~\underline{a}ttention-centric \underline{e}fficient and \underline{s}calable \underline{p}ost-training quantization \underline{a}lgorithm} that pursues both accuracy and efficiency.
The key idea of \aespa \ is to perform quantization layer-wise for efficiency while targeting the attention-wise reconstruction to consider the cross-layer dependency.

Our contributions are summarized as follows:
\vspace{-1.5mm}
\begin{itemize}[leftmargin=1em]
    \item We propose a new quantization strategy that balances accuracy and efficiency.
    Our scheme aims to reconstruct the attention output to consider the cross-layer dependency while quantizing models layer-wise to pursue efficiency.
    \item To accelerate the quantization process, we propose refined quantization objectives for the attention module.
    Through a complexity analysis, we demonstrate that quantization that is approximately 10 times faster than existing block-wise approaches can be achieved by exploiting the proposed objectives.
    \item From extensive experiments on language models, we demonstrate that our approach outperforms conventional schemes by a significant margin, particularly for low-bit precision (INT2).
\end{itemize}

\section{Background}

\subsection{Classic PTQ methods}

Recent studies on PTQ have mostly attempted to minimize the increase in the task loss incurred by quantization rather than the quantization error itself ($\Delta \mW$).
Consider a pre-trained neural network parameterized by weights $\mW$.
If we assume the well-convergence of the network, the problem of quantizing weights $\mW$ to minimize the loss degradation can be formulated as~\cite{lecun1989optimal, nagel2020up}
\begin{align}
    \min_{\Delta \vw}~~\mathbb{E} \left [ \Delta \vw^{T} \cdot \mathbf{H}^{(\vw)} \cdot \Delta \vw \right ],
    \label{eq:ptq_goal_refined}
\end{align}
where $\mathbf{H}^{(\vw)}$ is the Hessian related to the flattened weight $\vw$.
Because computing and storing $\mathbf{H}^{(\vw)}$ is infeasible, further assumptions have been made to simplify~\cref{eq:ptq_goal_refined}.
In~\cite{nagel2020up}, for example, layer-wise independence has been assumed, relaxing~\cref{eq:ptq_goal_refined} into the layer-wise reconstruction problem:
\begin{align}
    \min_{\Delta \mW^{(\ell)}}~& \mathbb{E} \left [ \left \| \mathcal{Q} ( \mW^{(\ell)} ) \mX - \mW^{(\ell)} \mX \right \|_{F}^{2} \right ], \label{eq:layer-wise reconstruction}
\end{align}
where $\mW^{(\ell)}$ denotes the weights of the $\ell$-th layer, $\mX$ is the input, and $\gQ$ is a quantization function. 
For a uniform quantization, if the nearest-rounding is used to assign integer weights, $\gQ$ is defined as
\begin{align}
    \mathcal{Q}(x)
        &= s \left ( \text{clamp} \left ( \left \lfloor \frac{x}{s} \right \rceil + z, 0, 2^{n}-1 \right ) - z \right ),\label{eq:quantizer}
\end{align}
where $s$, $z$, and $n$ are the scale, zero-point, and bit-width, respectively, and $\lfloor \cdot \rceil$ represents the round-off.

Early studies on PTQ focused on optimizing the weight-rounding policy~\cite{nagel2020up, hubara2021accurate, li2021brecq, jeon2022mr, jeon2023genie}.
These studies have attempted to assign each weight to a ``proper'' grid (instead of an adjacent grid), such that the loss degradation could be minimized.
In~\cite{nagel2020up}, a learning-based weight-rounding optimization algorithm, called AdaRound, has been proposed to solve the layer-wise reconstruction problem in~\cref{eq:layer-wise reconstruction}.
In~\cite{li2021brecq}, AdaRound has been extended to the following block-wise reconstruction problem:
\begin{align}
    \min_{\Delta \mW^{(\ell)}}~& \mathbb{E} \left [ \left \| f^{(\ell)} \left ( \mathcal{Q} ( \mW^{(\ell)} ), \mX \right ) - f^{(\ell)} \left ( \mW^{(\ell)}, \mX \right ) \right \|_{F}^{2} \right ], \label{eq:block-wise reconstruction}
\end{align}
where $\mW^{(\ell)}$ denotes the weights of the $\ell$-th block $f^{(\ell)}$ (\eg \ ResNet or Transformer block).
By considering the dependency between layers inside the block, this algorithm, termed \brecq, not only performs better than AdaRound, but also exhibits robust performance for a low bit-width (\eg \ INT2).

\subsection{PTQ for LLMs}

Although AdaRound and \brecq~have been successful in quantizing small-scale networks (\eg \ ResNet), scaling those learning-based schemes to LLMs with billions of parameters is challenging.
In fact, \brecq~requires more than 20 GPU hours to quantize relatively small-sized language models (\eg \ OPT-2.7B; see \cref{appendix:time-memory-costs}), which would not be suitable for the real-world deployment of LLMs where models to be deployed are frequently updated.

Owing to the excessive time and memory costs of classic PTQ schemes, recent studies have focused on developing cost-effective alternatives for quantizing LLMs.
In OPTQ~\cite{frantar2023optq}, a one-shot PTQ scheme that optimizes a weight-rounding policy without relying on learning, has been proposed.
In addition, PTQ schemes that enhance the performance of the nearest-rounding, rather than optimizing the weight-rounding policy, have been proposed.
These schemes use additional ``foldable'' parameters\footnote{By foldable parameters, we mean the parameters that can be merged into other layers within the Transformer block (\eg \ LayerNorm), thereby imposing no extra computational cost during the inference~\cite{jeon2023frustratingly}.} to suppress activation outliers or quantize weights more precisely~\cite{xiao2023smoothquant, lin2023awq, jeon2023frustratingly, shao2023omniquant, ma2024affinequant}. 

Although previous studies have mitigated the computational overhead of classic PTQ methods, they often sacrifice the low-bit quantization performance or suffer from an unstable quantization process.
The main reason for this unsatisfactory performance is that all the schemes mentioned above, except OPTQ, rely on nearest-rounding and do not optimize the weight-rounding policy. 
Moreover, most of them target layer-wise reconstruction in~\cref{eq:layer-wise reconstruction}, not block-wise reconstruction in~\cref{eq:block-wise reconstruction}, thus ignoring the cross-layer dependency within the attention module.
Although~\cite{shao2023omniquant, ma2024affinequant} target block-wise reconstruction via learning, they need to approximate gradients for a non-differentiable quantization function, which results in an unstable training process (see \cref{tab:comparison-with-block-wise-methods} in \cref{sec:experimental results})~\cite{lin2023awq}.

Thus, we propose a novel PTQ scheme that balances accuracy and efficiency.
In contrast to conventional LLM quantization methods, our scheme optimizes a weight-rounding policy while targeting block-wise reconstruction to consider the cross-layer dependency.
The key difference over classic block-wise weight-rounding optimization is that we quantize models layer-wise for scalability, whereas layers are jointly quantized in the existing methods.
Furthermore, we present an efficient pre-computation-based method for the computation of the block-wise objective in~\cref{eq:block-wise reconstruction}, which significantly reduces the computational overhead caused by repeated attention operations.

\section{Method} \label{sec:method}

\subsection{Motivation}

\begin{figure*}[t]
    \centering
    \includegraphics[width=.85\linewidth]{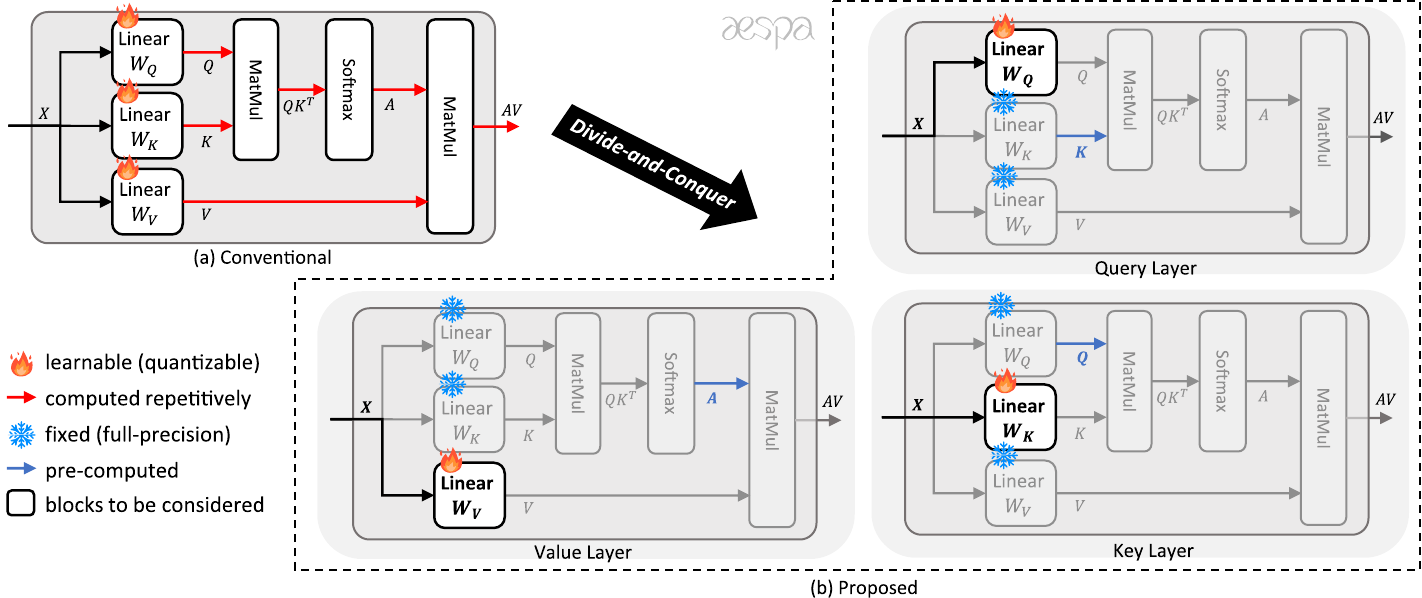}
    \vspace{-0.2cm}
    \caption{Overview of \aespa. Each weight is quantized separately to reconstruct the attention output.}
    \vspace{-0.2cm}
    \label{fig:all}
\end{figure*}

To gain insight into our approach, we first consider the objective of the layer-wise reconstruction in~\cref{eq:layer-wise reconstruction}.
Let $\Delta \mW^{(\ell)} = \mathcal{Q}(\mW^{(\ell)}) - \mW^{(\ell)}$, then the reconstruction error can be expressed as 
\begin{align}
    \mathbb{E} \left [ \left \| \Delta \mW \mX \right \|_{F}^{2} \right ]
        =\hspace{-.5mm} \mathbb{E} \left [ \trace \hspace{-.5mm} \left ( \Delta \mW \mX \mX^{T} \Delta \mW^{T} \right ) \right ]
        =\hspace{-.5mm} \trace \hspace{-.5mm} \left ( \Delta \mW \hspace{-.5mm} \cdot \hspace{-.5mm} \mathbb{E} \left [ \mX \mX^{T} \right ] \hspace{-.5mm} \cdot \hspace{-.5mm} \Delta \mW^{T} \right ) \hspace{-.5mm}. \label{eq:asepa_layer-wise recon error}
\end{align}
Consequently, the layer-wise quantization problem can be recast as follows:
\begin{align}
    \min_{\Delta \mW}~\trace \left ( \Delta \mW \cdot \mathbb{E} \left [ \mX \mX^{T} \right ] \cdot \Delta \mW^{T} \right ). \label{eq:aespa_layer-wise quantization}
\end{align}
The new form of the quantization objective in~\cref{eq:aespa_layer-wise quantization} implies that if $\mathbb{E}[\mX \mX^{T}]$ is pre-computed and stored before quantization, we can measure the reconstruction error over the entire calibration dataset with a single matrix multiplication and element-wise multiplication.\footnote{We note that the computation of $\trace (\mathbf{A} \mathbf{B} \mathbf{C}^{T})$ can be implemented as $\texttt{torch.sum}((\mathbf{A} \mathbf{B}) \odot \mathbf{C})$, where $\odot$ denotes the element-wise product operation. They are mathematically equivalent. \label{footnote:trace}}
This is in contrast to the original formulation in~\cref{eq:layer-wise reconstruction} which requires the computation of $\mathcal{Q}(\mW)\mX$ or $\Delta \mW \mX$ for every input $\mX$.

A natural question that arises from this finding is \textit{``Can we also measure the block reconstruction error efficiently based on such a pre-computation?''}.
In the following subsections, we describe our main strategy to simplify block-wise quantization and then present a refined objective for the attention module, where the objective can be computed efficiently with certain pre-computed values.

\begin{figure*}[t]
  \begin{subfigure}{0.33\linewidth}
    \centering
    \includegraphics[width=.8\textwidth, keepaspectratio]{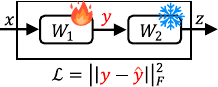}
    \vspace{-0.2cm}
    \caption{\fontsize{7.5pt}{9.0pt}\selectfont \textbf{Layer}-Wise, \textbf{Layer} Output}
    \label{fig:adaround}
  \end{subfigure}
  \begin{subfigure}{0.33\linewidth}
    \centering
    \includegraphics[width=.8\textwidth, keepaspectratio]{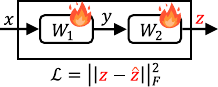}
    \vspace{-0.2cm}
    \caption{\fontsize{7.5pt}{9.0pt}\selectfont\textbf{Block}-Wise, \textbf{Block} Output}
    \label{fig:brecq}
  \end{subfigure}
  \begin{subfigure}{0.33\linewidth}
    \centering
    \includegraphics[width=.8\textwidth, keepaspectratio]{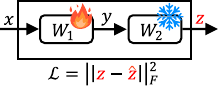}
    \vspace{-0.2cm}
    \caption{\fontsize{7.5pt}{9.0pt}\selectfont\textbf{Layer}-Wise, \textbf{Block} Output (Proposed)}
    \label{fig:hybird}
  \end{subfigure}
  \caption{Quantization strategies (simplified)}
  \vspace{-0.4cm}
  \label{fig:qled}
\end{figure*}

\subsection{Quantization strategy of \aespa}  \label{subsec:qled}

When quantizing the attention module using conventional block-wise reconstruction methods (\cref{fig:all}(a)), the query, key, and value projections have been jointly optimized such that
\begin{align} \label{eq:attn recon error}
    \min_{\Delta \mW_{Q}, \Delta \mW_{K}, \Delta \mW_{V}}~\mathbb{E} \hspace{-.5mm} \left [ \hspace{-.5mm} \left \| \SA (\widehat{\mQ}, \widehat{\mK}, \widehat{\mV}) \hspace{-.5mm} - \hspace{-.5mm} \SA(\mQ, \mK, \mV) \right \|_{F}^{2} \hspace{-.5mm} \right ],
\end{align}
where the output of attention module $\SA(\mQ, \mK, \mV)$ is defined as 
\begin{align} 
    \SA(\mQ, \mK, \mV) 
        = \softmax \left ( \frac{\mQ \mK^{T}}{\sqrt{d}} \right ) \mV 
        = \mA \mV.
\end{align}
In such a case, we need to compute $\SA (\widehat{\mQ}, \widehat{\mK}, \widehat{\mV})$ for every batch sequence in each iteration, which is computationally heavy and time-consuming (see \cref{subsec:complexity} for details on complexity).

To overcome this computational overhead, we quantize each projection \textit{separately} in a divide-and-conquer manner.
For example, when quantizing the query projection $\mW_{Q}$, we fix $\mW_{K}$ and $\mW_{V}$ with full-precision (\cref{fig:all}(b)), which facilitates the factoring out of common terms affected by $\mW_{K}$ and $\mW_{V}$ (see~\cref{subsec:refined objectives} for details).
We emphasize that this strategy differs from conventional layer-wise quantization schemes (\eg \ AdaRound and OPTQ) in that we aim to minimize the reconstruction error for the attention module, not the reconstruction error for each layer.

We conduct experiments to demonstrate the importance of targeting attention-wise reconstruction and validity of the proposed quantization strategy.
In our experiments, we set the loss function for each projection as the attention reconstruction error in~\cref{eq:attn recon error} but quantize each projection separately (see \cref{fig:qled}(c)).
\cref{tab:qled_poc} in \cref{appendix: validity-of-qled} summarizes the performance of AdaRound, \brecq, and our approach.
As evident, our approach uniformly outperforms AdaRound for all bit-widths, although both methods quantize models layer-wise.
This is because we can consider cross-layer dependency (\ie \ relationship between the query, key, and value) by targeting attention-wise reconstruction, which is different from AdaRound wherein layers are considered independent.
Furthermore, once we target attention-wise reconstruction, separate layer-wise quantization does not incur severe performance degradation compared to the joint quantization method (\brecq).
In fact, our approach causes only a marginal performance degradation for 2-bit and exhibits comparable performance for 3-bit and 4-bit.
For further discussion on the proposed strategy, see \cref{appendix: validity-of-qled}.

\subsection{Refined quantization objectives for \aespa} \label{subsec:refined objectives}

One might ask whether our strategy incurs more computational cost than that required by the joint quantization because we update only one layer at a time (see \cref{fig:all}(b)).
This is in contrast to existing methods, in which the layers inside the attention module are updated simultaneously (\cref{fig:all}(a)).
To reduce this additional cost, we refine the quantization objective in~\cref{eq:attn recon error} for each projection. 

\textbf{Value projection}
When quantizing the value projection $\mW_{V}$, the query and key projections are fixed with full-precision.
In this case, by factoring out the common term influenced by $\mQ$ and $\mK$, we can simplify the attention reconstruction error $\Delta \SA_{V}$ as follows:
\begin{align}
    \Delta \SA_{V}
        &= \mathbb{E} \left [ \left \| \mA \widehat{\mV} - \mA \mV \right \|_{F}^{2} \right ] 
         = \mathbb{E} \left [ \left \| \mA \Delta \mV \right \|_{F}^{2} \right ] 
         = \mathbb{E} \left [ \left \| \Delta \mW_{V} \mX \mA^{T} \right \|_{F}^{2} \right ].
\end{align}
Thus, the problem to quantize $\mW_{V}$ to minimize the attention reconstruction error can be recast as
\begin{align}
    \min_{\Delta \mW_{V}}~\mathbb{E} \left [ \left \| \Delta \mW_{V} \mX \mA^{T} \right \|_{F}^{2} \right ]. \label{eq:quant goal_initial_V}
\end{align}

\textbf{Query projection}
When the key and value projections are fixed with full-precision, the attention reconstruction error $\Delta \SA_{Q}$ caused by $\Delta \mW_{Q}$ is expressed as
\begin{align} \label{eq:recon error_initial_Q} 
    \Delta \SA_{Q}
        &= \mathbb{E} \left [ \left \| \SA(\widehat{\mQ}, \mK, \mV) - \SA(\mQ, \mK, \mV) \right \|_{F}^{2} \right ] 
        = \mathbb{E} \left [ \left \| \Delta \mA \mV \right \|_{F}^{2} \right ],
\end{align}
where $\Delta \mA = \softmax ( \widehat{\mQ} \mK^{T} / \sqrt{d} ) - \softmax ( \mQ \mK^{T} / \sqrt{d})$. 
To avoid the computational overhead of repetitive softmax operations, we approximate $\Delta \mA$ with its first-order Taylor series as
\begin{align}
    \Delta \mA
        &\approx \frac{\Delta \mQ \mK^{T}}{\sqrt{d}} \cdot \mathbf{J}_{\softmax}^{T}, \label{eq:softmax approximation}
\end{align}
where $\mathbf{J}_{\softmax}$ is the Jacobian of the softmax function.
By combining~\cref{eq:recon error_initial_Q} and~\cref{eq:softmax approximation}, we obtain
\begin{align}
    \Delta \SA_{Q}
        &\approx \frac{1}{d} \mathbb{E} \left [ \left \| \Delta \mQ \mK^{T} \mathbf{J}_{\softmax}^{T} \mV \right \|_{F}^{2} \right ] 
        = \frac{1}{d} \mathbb{E} \left [ \left \| \mV^{T} \mathbf{J}_{\softmax} \mK \Delta \mW_{Q} \mX \right \|_{F}^{2} \right ]. \label{eq:recon error_mid_Q}
\end{align}
Although we can circumvent conducting attention operations using the modified form in~\cref{eq:recon error_mid_Q}, a large amount of memory is required to store the Jacobian $\mathbf{J}_{\softmax}$ (\eg \ more than 100~GB of memory for OPT-125M).\footnote{Note that the shape of $\mathbf{J}_{\softmax}$ is $[L, L, L]$ ($L$ is the input sequence length) for each attention head because $\mathbf{J}_{\softmax}(\va_{\ell}) = \diag(\va_{\ell}) - \va_{\ell}^{T} \va_{\ell} \in \mathbb{R}^{L \times L}$ for each row $\va_{\ell}$ of $\mA$. \label{footnote:jacobian_shape}}
As a cost-effective alternative, we build an upper bound of~\cref{eq:recon error_mid_Q} and then employ it as a surrogate of $\Delta \SA_{Q}$ when quantizing $\mW_{Q}$.
Specifically, by noting that 
\begin{align}
    \left \| \mV^{T} \mathbf{J}_{\softmax} \mK \Delta \mW_{Q} \mX \right \|_{F}^{2}
        \le \left \| \mV^{T} \mathbf{J}_{\softmax} \right \|_{F}^{2} \cdot \left \| \mK \Delta \mW_{Q} \mX \right \|_{F}^{2}
\end{align}
and the term $\| \mV^{T} \mathbf{J}_{\softmax} \|_{F}^{2}$ is fixed in the quantization process, we minimize $\| \mK \Delta \mW_{Q} \mX \|_{F}^{2}$ with the hope that $\Delta \SA_{Q}$ also decreases.
In other words, our quantization objective for $\mW_{Q}$ is
\begin{align}
    \min_{\Delta \mW_{Q}}~\mathbb{E} \left [ \left \| \mK \Delta \mW_{Q} \mX \right \|_{F}^{2} \right ]. \label{eq:quant goal_initial_Q}
\end{align}

\textbf{Key projection}
By taking similar steps, the quantization objective for the key projection $\mW_{K}$ can be formulated as (see \cref{appendix: proof_K} for the detailed derivation)
\begin{align} \label{eq:quant goal_initial_K}
    \min_{\Delta \mW_{K}}~\mathbb{E} \left [ \left \| \mQ \Delta \mW_{K} \mX \right \|_{F}^{2} \right ]. 
\end{align}

\subsection{Algorithm description}  \label{subsec:algo}

The proposed \aespa~consists of two main steps.
Specifically, \aespa~first determines the quantization parameters (\ie \ scale and zero-point) and then optimizes an integer weight $\mW_{int}$ for each weight.

Note that we only used the definition of the attention operation when developing the refined objectives in~\cref{eq:quant goal_initial_V},~\cref{eq:quant goal_initial_Q}, and~\cref{eq:quant goal_initial_K}.
Thus, our objectives can be integrated into any layer-wise quantization scheme without effort.
For example, we can compute the quantization parameters by combining existing parameter initialization algorithms (\eg \ AWQ~\cite{lin2023awq} and \zfold~\cite{jeon2023frustratingly}) with the proposed objectives.
We can also optimize a weight-rounding policy using conventional methods (\eg \ AdaRound~\cite{nagel2020up}) together with our objectives (see \cref{appendix: integration_adaround} for details).
In the proposed \aespa, we use \zfold \ in computing the quantization parameters and employ AdaRound in optimizing a weight-rounding policy.
In \cref{algo:aespa} (see \cref{appendix:pseudo-code}), we summarize the proposed \aespa.\footnote{We use the layer-wise objective in~\cref{eq:aespa_layer-wise quantization} for the weights other than the query, key, and value projections (\ie \ out-projection and weights inside the feed-forward network).}

To accelerate the weight-rounding learning process, we further modify the objective functions such that the value can be computed efficiently via pre-computation, as in~\cref{eq:asepa_layer-wise recon error}.

\textbf{Modified objective for~\cref{eq:quant goal_initial_V}}
The proposed objective for the value projection can be recast as
\begin{align} \label{eq:aespa_final recon error_V}
    \mathbb{E} \left [ \left \| \Delta \mW_{V} \mX \mA^{T} \right \|_{F}^{2} \right ]
        = \trace \left ( \Delta \mW_{V} \mathbb{E} \left [ \mX \mA^{T} \mA \mX^{T} \right ] \Delta \mW_{V}^{T} \right ).
\end{align}
The modified objective allows us to perform each iteration of the weight-rounding learning efficiently.
Specifically, by computing $\mathbb{E} [ \mX \mA^{T} \mA \mX^{T} ]$ before quantization and reusing it in the quantization process\footnote{The term $\mathbb{E} [ \mX \mA^{T} \mA \mX^{T} ]$ is not affected by $\Delta \mW_{V}$ and thus fixed in the quantization process.}, we can avoid the overhead of computing $\left \| \Delta \mW_{V} \mX \mA^{T} \right \|_{F}^{2}$ for every input $\mX$ and compute the loss with one simple matrix multiplication and a single element-wise multiplication (see \cref{footnote:trace}).

Another intriguing feature of this modification is that it facilitates a more reliable update of $\Delta \mW_{V}$ than the original objective in~\cref{eq:quant goal_initial_V}.
Specifically, because $\mathbb{E} [ \mX \mA^{T} \mA \mX^{T} ]$ is pre-computed using all calibration data, the loss computed with~\cref{eq:aespa_final recon error_V} considers the entire calibration dataset (\ie \ the batch size is the total number of data).
Thus, a better estimate of the true gradient can be obtained without any memory issues, which could lead to more consistent updates of $\Delta \mW_{V}$ and faster convergence~\cite{smith2018don}.

The modified objective in~\cref{eq:aespa_final recon error_V} also implies that the Hessian $\mathbf{H}_{V}$ for each row of $\mW_{V}$ is
\begin{align}
    \mathbf{H}_{V} 
        &= 2\mathbb{E} [ \mX \mA^{T} \mA \mX^{T} ]. \label{eq:aespa_Hessian_V}
\end{align}
We note that the proposed Hessian $\mathbf{H}_{V}$ differs from $\mathbf{H} = 2\mathbb{E} [\mX \mX^{T}]$, which has been commonly used as an approximated Hessian in conventional methods~\cite{obq, frantar2023optq, jeon2023frustratingly, chee2023quip}.
The key reason for the difference is that we consider the dependency between $\mW_{Q}$, $\mW_{K}$, and $\mW_{V}$ by targeting attention-wise reconstruction, whereas the previous methods assumed independence.
To observe the effect of considering the cross-layer dependency, we use different Hessians (\ie \ $\mathbf{H}_{V}$ and $\mathbf{H}$) when quantizing language models and then compare the performance of the quantized models (see \cref{appendix:hessian_ablation}). 
Evidently, the quantization performance is much better when the proposed Hessian $\mathbf{H}_{V}$ is used, which demonstrates the importance of considering the cross-layer dependency.

\textbf{Modified objectives for~\cref{eq:quant goal_initial_Q} and~\cref{eq:quant goal_initial_K}}
If we denote the vectorized representation of $\Delta \mW_{Q}$ as $\Delta \vw_{Q}$, the proposed objective in~\cref{eq:quant goal_initial_Q} can be expressed as (see \cref{appendix: proof_Q} for the derivation)
\begin{align}
    \mathbb{E} \hspace{-.5mm} \left [ \left \| \mK \Delta \mW_{Q} \mX \right \|_{F}^{2} \right ]
        &\hspace{-.5mm}=\hspace{-.5mm} \Delta \vw_{Q}^{T} \cdot \mathbb{E} \hspace{-.5mm} \left [ \mX \mX^{T} \hspace{-.5mm} \otimes \hspace{-.5mm} \mK^{T} \mK \right ] \cdot \Delta \vw_{Q}. \label{eq:quant goal_mid_Q}
\end{align}
where $\otimes$ is the Kronecker product operation.
To reduce the memory cost of storing the Kronecker product term $\mathbb{E} \hspace{-.5mm} \left [ \mX \mX^{T} \hspace{-.5mm} \otimes \hspace{-.5mm} \mK^{T} \mK \right ]$, we approximate it as~\cite{botev2017practical}
\begin{align} \label{eq:kronecker approximation}
    \mathbb{E} \left [ \mX \mX^{T} \otimes \mK^{T} \mK \right ]
        &\approx \mathbb{E} \left [ \mX \mX^{T} \right ] \otimes \mathbb{E} \left [ \mK^{T} \mK \right ].
\end{align}
By combining~\cref{eq:quant goal_mid_Q} and~\cref{eq:kronecker approximation}, we obtain
\begin{align}
    \mathbb{E} \hspace{-.5mm} \left [ \left \| \mK \Delta \mW_{Q} \mX \right \|_{F}^{2} \right ]
        &\approx \Delta \vw_{Q}^{T} \cdot \left ( \mathbb{E} \left [ \mX \mX^{T} \right ] \otimes \mathbb{E} \left [ \mK^{T} \mK \right ] \right ) \cdot \Delta \vw_{Q} \nonumber \\
        &\hspace{-.37mm}\overset{(a)}{=} \trace \hspace{-.5mm} \left ( \mathbb{E} \hspace{-.5mm} \left [ \mK^{T} \mK \right ] \hspace{-.5mm} \Delta \mW_{Q} \mathbb{E} \left [ \mX \mX^{T} \right ] \hspace{-.5mm} \Delta \mW_{Q}^{T} \right ),
        \label{eq:quant goal_final_Q}
\end{align}
where the proof of (a) is provided in \cref{appendix: proof_Q}.
By taking similar steps, the objective for the key projection can be recast as
\begin{align}
    \mathbb{E} \left [ \left \| \mQ \Delta \mW_{K} \mX \right \|_{F}^{2} \right ]
        &= \trace \hspace{-.5mm} \left ( \mathbb{E} \hspace{-.5mm} \left [ \mQ^{T} \mQ \right ] \hspace{-.5mm} \Delta \mW_{K} \mathbb{E} \left [ \mX \mX^{T} \right ] \hspace{-.5mm} \Delta \mW_{K}^{T} \right ). \label{eq:quant goal_final_K}
\end{align}
The modified objectives in~\cref{eq:quant goal_final_Q} and~\cref{eq:quant goal_final_K} imply that the loss over the total calibration dataset can be calculated efficiently by computing $\mathbb{E}[\mK^{T} \mK]$, $\mathbb{E} [\mQ^{T} \mQ]$, and $\mathbb{E} [\mX \mX^{T}]$ in advance.

\subsection{Complexity analysis for \aespa}  \label{subsec:complexity}

We discuss the computational complexity of \aespa. 
Specifically, we analyze the number of floating-point operations (flops) required to perform one iteration for weight-rounding optimization (line~6 in \cref{algo:aespa}).
For each projection, the required number of flops is summarized as follows.
\begin{itemize} [leftmargin=1.5em]
    \item {\bf{Value}}: By reusing the pre-computed $\mathbb{E} [\mX \mA^{T} \mA \mX^{T}]$, the loss value in~\cref{eq:aespa_final recon error_V} can be computed with one matrix multiplication and one element-wise multiplication/addition (see \cref{footnote:trace}).
    The associated cost is $2d_{h}d^{2} + d_{h}d - 1$ flops, where $d$ is the hidden size and $d_{h}$ is the input dimension for each attention head.
    
    \item {\bf{Query/key}}: Once $\mathbb{E}[\mK^{T} \mK]$, $\mathbb{E}[\mQ^{T} \mQ]$, and $\mathbb{E}[\mX \mX^{T}]$ have been computed in advance, the loss values in~\cref{eq:quant goal_final_Q} and~\cref{eq:quant goal_final_K} can be computed by performing two matrix multiplications and one element-wise multiplication/addition.
    This requires $2d_{h}d^{2} +2d_{h}^{2}d - 1$ flops for each projection.
\end{itemize}
To summarize, the total number of flops required in each iteration of the proposed \aespa \ is
\begin{align}  \label{eq:cost_aespa}
    \gC_{\aespa} 
        &= 6d_{h}d^{2} + 4d_{h}^{2}d + d_{h}d - 3
        = \gO(d_{h}d^{2}).
\end{align}
We emphasize that regardless of the amount of calibration data, the number of flops to compute the loss considering the entire dataset is fixed as $\gC_{\aespa}$.

We now compare the complexities of \aespa \ and conventional block-wise quantization methods.
It can be easily verified that the existing methods require the following number of flops for handling $B$ input sequences of length $L$ (see \cref{appendix:complexity_conventional}):
\begin{align}  \label{eq:cost_conventional}
    \gC_{exist} 
        = B(6d_{h}dL + 4d_{h}L^{2} +2L^{2} - L - 1) 
        = \gO(Bd_{h}L \cdot \max \{ d, L \}).
\end{align}
\cref{tab:complexity} in \cref{appendix:complexity_conventional} summarizes the computational costs for different sizes of OPT models.
For the conventional methods, we report the cost of using four sequences in each iteration ($B=4$).
We observe that the computational cost of \aespa \ is considerably lower than that of conventional methods.
In particular, for small-scale models, \aespa \ performs ten times fewer number of flops.
It can be observed that the gap between $\gC_{\aespa}$ and $\gC_{exist}$ decreases as the model size increases.
This is because the hidden size $d$ exceeds the sequence length $L$ (which is fixed for all models) for large models.
Nevertheless, \aespa \ still incurs a lower computational cost, and the gap increases if conventional methods use larger batch sizes.

\section{Experimental results} \label{sec:experimental results}

\subsection{Experimental setup}  \label{subsec:setup}

We quantize publicly available LLMs (\eg \ OPT~\cite{zhang2022opt}, BLOOM~\cite{scao2022bloom}, LLaMA~\cite{touvron2023llama}, and LLaMA2~\cite{touvron2023llama2}) using the proposed \aespa.
When implementing \aespa, we compute the quantization parameters with \zfold~\cite{jeon2023frustratingly} and optimize a weight-rounding policy via AdaRound~\cite{nagel2020up}, where the proposed row-wise Hessians and loss functions (see~\cref{tab:algo} in \cref{appendix:pseudo-code}) are utilized instead of the existing ones.
When computing the quantization parameters, we follow the stopping criterion introduced by~\cite{jeon2023frustratingly}.
Before optimizing a weight-rounding policy, we update the full-precision weights via OPTQ~\cite{frantar2023optq}, which empirically reduces the number of iterations required for weight-rounding optimization.
When optimizing a weight-rounding policy, we set the number of iterations, learning rate, and weight of the rounding loss (see $\lambda$ in~\cref{eq:aespa_adaround}) to 2,000, 0.015, and 1.5, respectively.

When constructing the calibration dataset, we randomly sample 128 segments consisting of 2048 tokens from the C4 dataset~\cite{c4} as in~\cite{frantar2023optq, jeon2023frustratingly, chee2023quip}.
In our experiments, we quantize only weights and retain activations in full-precision because activations are not a significant bottleneck for LLMs~\cite{frantar2023optq} and the inference of LLMs can be accelerated sufficiently by reducing memory movement through weight quantization~\cite{kim2023squeezellm}.
We evaluate the performance of the quantized models using benchmark datasets (\eg \ WikiText-2~\cite{wiki}, C4~\cite{c4}, and PTB~\cite{ptb}) and zero-shot tasks.
Except for the experiments on the LLaMA2 models, which were performed using an NVIDIA H100 GPU, we conducted all experiments using a single NVIDIA A100 GPU (80 GB).

\subsection{Comparison with prior arts}

\begin{table*}[t]
    \renewcommand{\arraystretch}{1.0}
    \fontsize{7.5pt}{9.0pt}\selectfont
    \centering
    \caption{Performance (PPL $\downarrow$) of the proposed \aespa~and conventional block-wise PTQ methods.}

    \vspace{-.1cm}
    
    \begin{subtable}{\textwidth}
        \centering
        \caption{WikiText-2}
        \vspace{-.2cm}
        \begin{tabular}{c l c c c c c c c c c c c}
        \toprule
        \multirow{2}{*}{Precision} & \multirow{2}{*}{Method} & \multicolumn{4}{c}{OPT} & & \multicolumn{3}{c}{LLaMA} & & \multicolumn{2}{c}{LLaMA2}\\
        \cline{3-6} \cline{8-10} \cline{12-13}
        & & 125M & 1.3B & 2.7B & 6.7B & & 7B & 13B & 30B & & 7B & 13B \\
        \toprule
        FP16 & Baseline & 27.65 & 14.63 & 12.47 & 10.86 &   & 5.677 & 5.091 & 4.101 & & 5.472 & 4.884 \\
        \midrule
        \multirow{4}{*}{INT3}
        & \brecq~\cite{li2021brecq}             & 33.25 & 16.09 & 13.37 & OOM & & OOM & OOM & OOM & & OOM & OOM \\
        & OmniQuant~\cite{shao2023omniquant}    & 39.14 & 17.59 & 14.87 & 12.87 & & 6.716 & 5.798 & 4.963 & & 6.798 & 5.751 \\
        & AffineQuant~\cite{ma2024affinequant}  & 36.15 & 17.26 & 14.25 & 12.30 & & 6.712 & 5.820 & 4.951 & & 6.795 & 5.757 \\
        & \bf{\small \aespa}                    & \bf{32.71} & \bf{15.79} & \bf{13.14} & \bf{11.23} & & \bf{6.579} & \bf{5.611} & \bf{4.688} & & \bf{6.241} & \bf{5.462} \\
        \midrule
        \multirow{4}{*}{INT2}
        & \brecq~\cite{li2021brecq}             & \bf{60.38} & 56.25 & 113.6 & OOM & & OOM & OOM & OOM & & OOM & OOM \\
        & OmniQuant~\cite{shao2023omniquant}    & NaN & 399.6 & 1.6e3 & 4.9e3 & & 18.18 & NaN & 10.15 & & 35.40 & 20.19 \\
        & AffineQuant~\cite{ma2024affinequant}  & 143.9 & 56.45 & 35.16 & 25.32 & & 18.83 & 11.08 & NaN & & NaN & 18.49 \\
        & \bf{\small\aespa}                     & 71.18 & \bf{24.26} & \bf{22.22} & \bf{15.71} & & \bf{11.94} & \bf{10.30} & \bf{7.845} & & \bf{13.99} & \bf{12.14} \\
        \bottomrule
        \end{tabular}
    \end{subtable}

    \vspace{.2cm}
    
    \begin{subtable}{\textwidth}
        \centering
        \caption{C4}
        \vspace{-.2cm}
        \begin{threeparttable}
        \begin{tabular}{c l c c c c c c c c c c c}
        \toprule
        \multirow{2}{*}{Precision} & \multirow{2}{*}{Method} & \multicolumn{4}{c}{OPT} & & \multicolumn{3}{c}{LLaMA} & & \multicolumn{2}{c}{LLaMA2}\\
        \cline{3-6} \cline{8-10} \cline{12-13}
        & & 125M & 1.3B & 2.7B & 6.7B & & 7B & 13B & 30B & & 7B & 13B \\
        \toprule
        FP16 & Baseline & 26.56 & 16.07 & 14.34 & 12.71 &   & 7.344  & 6.798 & 6.131 & & 7.264 & 6.727 \\
        \midrule
        \multirow{4}{*}{INT3}
        & \brecq~\cite{li2021brecq}               & 29.74 & 17.46 & 15.39 & OOM & & OOM & OOM & OOM & & OOM & OOM \\
        & OmniQuant~\cite{shao2023omniquant}      & 34.92 & 18.83 & 16.80 & 14.21 & & 8.605 & 7.604 & 6.822 & & 9.085 & 7.821 \\
        & AffineQuant~\cite{ma2024affinequant}    & 32.78 & 18.27 & 16.11 & 13.80 & & 8.631 & 7.609 & 6.803 & & 9.059 & 7.732 \\
        & \bf{\small\aespa}                       & \bf{29.51} & \bf{17.10} & \bf{15.27} & \bf{13.15} & & \bf{8.465} & \bf{7.399} & \bf{6.634} & & \bf{8.225} & \bf{7.392} \\
        \midrule
        \multirow{4}{*}{INT2}
        & \brecq~\cite{li2021brecq}               & \bf{47.85} & 41.05 & 83.32 & OOM & & OOM & OOM & OOM & & OOM & OOM \\
        & OmniQuant~\cite{shao2023omniquant}      & NaN & 239.1 & 1.1e3 & 4.4e3 & & 18.59 & NaN & 14.74 & & 26.27 & 18.93 \\
        & AffineQuant~\cite{ma2024affinequant}    & 95.86 & 43.66 & 29.75 & 24.04 & & 16.87 & 12.79 & NaN & & NaN & 15.20 \\
        & \bf{\small\aespa}                       & 56.88 & \bf{23.54} & \bf{22.53} & \bf{17.28} & & \bf{13.63} & \bf{11.46} & \bf{10.35} & & \bf{14.36} & \bf{13.59} \\
        \bottomrule
        \end{tabular}
        \begin{tablenotes}
            \item[*] `NaN' means that loss diverges in the quantization process.
            \item[*] `OOM' means that out-of-memory issues occur when quantizing models with a single A100 GPU.
            \item[*] Results for high bit-widths are provided in~\cref{appendix:comparison with block-wise schemes_PTB} due to the page limitation.
        \end{tablenotes}
        \end{threeparttable}
    \end{subtable}
    
    \label{tab:comparison-with-block-wise-methods}

    \vspace{-.3cm}
    
\end{table*}

\textbf{Comparison with block-wise PTQ schemes}
We compare the proposed \aespa \ with conventional block-wise PTQ methods, among which \brecq \ is a classic weight-rounding optimization method, and OmniQuant and AffineQuant are LLM quantization methods that mitigate the computational overhead of \brecq \ by learning only a few quantization and foldable parameters~\cite{shao2023omniquant, ma2024affinequant}.
For OmniQuant and AffineQuant, we ran the official codes\footnote{https://github.com/OpenGVLab/OmniQuant, https://github.com/bytedance/AffineQuant \label{footnote:codes}}  provided by the authors.
For both methods, we activated the learnable equivalent transformation (LET) and learnable weight clipping (LWC) options and reported the obtained results.
When implementing \brecq, we employed the hyperparameter settings provided in~\cite{li2021brecq}.
In this comparison, the BLOOM models and OPT-350M were excluded because they are not supported by OmniQuant and AffineQuant.

As \cref{tab:comparison-with-block-wise-methods} shows, \aespa \ uniformly outperforms OmniQuant/AffineQuant.\footnote{We note that our results are different from those reported in~\cite{shao2023omniquant, ma2024affinequant} where a different calibration dataset (WikiText-2) was used; see \cref{appendix:wiki2} for more discussion on this issue.}
In particular, the performance gap is significant for 2-bit; while OmniQuant/AffineQuant suffer from instability (\ie \ loss diverges) or collapse (perplexity (PPL) $> 10^{3}$), \aespa \ exhibits reasonable PPL.
The outstanding performance is attributed to the fact that \aespa \ optimizes a weight-rounding policy after determining the quantization parameters (lines 5-8 in \cref{algo:aespa}), whereas OmniQuant/AffineQuant rely on the naive nearest rounding and approximate gradients for the non-differentiable quantization function.

Although \brecq \ performs best for the 2-bit quantization of OPT-125M, it lacks scalability; \brecq \ requires approximately 20 GPU hours for a relatively small-scale OPT-2.7B (see \cref{tab:time-comparison} in \cref{appendix:time-memory-costs}).
Even for OPT-125M, \brecq \ requires approximately 2 GPU hours, whereas the proposed \aespa \ completes quantization in 5 minutes.
One might wonder why the performance of \brecq \ worsens as the model size increases.
We assume that this is attributable to the choice of hyperparameters (\eg \ learning rate and weight of rounding loss). 
In fact, the hyperparameters presented in~\cite{li2021brecq} have been optimized for ImageNet, but not for LLMs.
It is expected that we can obtain better performance for \brecq \ via deliberate hyperparameter tuning; however, this would not be feasible for real-world deployment because it requires considerable time (see \cref{tab:time-comparison} in \cref{appendix:time-memory-costs}).

\begin{table*}[hbt!]
    \renewcommand{\arraystretch}{1.0}
    \fontsize{7.5pt}{9.0pt}\selectfont
    \centering
    \caption{Performance (PPL $\downarrow$) of \aespa~and existing layer-wise PTQ methods on BLOOM models.}
    \vspace{-.2cm}
    \begin{threeparttable}
    \begin{tabular}{c l c c c c c c c c c c c}
    \toprule
    \multirowcell{2}{Precision} & \multirowcell{2}{Method} & \multicolumn{5}{c}{WikiText-2} & & \multicolumn{5}{c}{C4}\\
    \cline{3-7} \cline{9-13}
    & & 560M & 1.1B & 1.7B & 3B & 7.1B & & 560M & 1.1B & 1.7B & 3B & 7.1B \\
    \toprule
    FP16 & Baseline & 22.42 & 17.69 & 15.39 & 13.48 & 11.37 & & 26.60 & 22.05 & 19.49 & 17.49 & 15.20 \\
    \midrule
    \multirow{4}{*}{INT3}
    & RTN                     & 56.74 & 49.85 & 63.37 & 39.07 & 17.35 & & 66.99	& 60.41	& 113.6	& 79.84	& 22.54 \\
    & OPTQ                    & 31.55 & 23.84 & 20.06 & 17.13 & 13.56 & & 34.62	& 27.62	& 23.87	& 20.96	& 17.43 \\
    & \zfold                  & 26.52 & 20.99 & 17.39 & 15.11 & 12.26 & & 29.97 & 24.43 & 21.52 & 19.01 & 16.12 \\
    & \bf{\footnotesize\aespa}  & \bf{25.39}    & \bf{19.81}    & \bf{16.95}    & \bf{14.68}    & \bf{12.00} & & \bf{29.10}    & \bf{23.80}    & \bf{20.93}    & \bf{18.55}    & \bf{15.91}    \\
    \midrule
    \multirow{4}{*}{INT2}
    & RTN                     & 7.8e5 & 9.8e5 & 3.5e5 & 1.4e5 & 2.1e5 & & 1.4e6	& 2.1e6	& 2.7e5	& 9.2e4	& 1.3e5 \\
    & OPTQ                    & 1.7e3 & 1.9e3 & 1.4e3 & 796.5 & 194.2 & & 533.4	& 538.0 & 562.9	& 351.6	& 112.8 \\
    & \zfold                  & 65.45 & 44.50 & 35.69 & 27.40 & 18.87 & & 64.11 & 42.96 & 37.26 & 32.64 & 22.46 \\
    & \bf{\footnotesize\aespa}  & \bf{44.91}    & \bf{34.12}    & \bf{27.67}    & \bf{21.65}    & \bf{16.31}    & & \bf{45.04}    & \bf{35.12}    & \bf{29.95}    & \bf{25.04}    & \bf{20.00}    \\
    \bottomrule
    \end{tabular}
    \begin{tablenotes}
        \item[*] Results for high bit-widths and other language models (\eg \ OPT, LLaMA, and LLaMA2) are provided in~\cref{appendix:comparison with one-shot schemes}.
    \end{tablenotes}
    \end{threeparttable}
    \label{tab:comparison-with-layer-wise-methods_bloom}

    \vspace{-.3cm}
\end{table*}

\textbf{Comparison with layer-wise PTQ schemes}
We compare the proposed \aespa \ with conventional layer-wise PTQ schemes, among which RTN is the method that naively assigns the nearest grid, OPTQ is a backpropagation-free weight-rounding optimization algorithm~\cite{frantar2023optq}, and \zfold \ is the method exploiting additional foldable parameters to quantize weights more elaborately~\cite{jeon2023frustratingly}.
\cref{tab:comparison-with-layer-wise-methods_bloom} and Tables~\ref{tab:opt}-\ref{tab:llama2} (see \cref{appendix:comparison with one-shot schemes}) summarize the results for the OPT, BLOOM, LLaMA, and LLaMA2 models of various sizes.
Evidently, \aespa \ uniformly outperforms conventional schemes, regardless of the size and type of LLMs.
In particular, for 2-bit, there exists a significant performance gap between \aespa \ and existing methods; the PPL obtained by \aespa \ is twice as low as those of conventional methods for small-scale models (\eg \ OPT-125M).
The key factors leading to such an outstanding performance are: 1) the consideration of the cross-layer dependency achieved by targeting attention-wise reconstruction, and 2) efficient weight-rounding optimization based on pre-computations.

\begin{table*}[t]
    \renewcommand{\arraystretch}{1.0}
    \fontsize{7.5pt}{9.0pt}\selectfont
    \centering
    \caption{INT2 zero-shot performance (accuracy $\uparrow$) of \aespa \ and existing methods.}
    \vspace{-.2cm}
    \begin{threeparttable}
    \begin{tabular}{c l c c c c c}
    \toprule
    Model & Method & ARC-c & ARC-e & HellaSwag & MMLU & Average \\
    \toprule
    \multirowcell{7.5}{LLaMA-7B} & FP16 & 44.62 & 72.85 & 76.18 & 32.19 & 56.46 \\
    \cmidrule{2-7}
    & RTN & 28.67 & 25.00 & 26.43 & 25.72 & 26.46 \\
    & OPTQ~\cite{frantar2023optq}             & 29.18 & 26.14 & 26.18 & 24.04 & 26.39 \\
    & \zfold~\cite{jeon2023frustratingly}     & 30.63 & 52.44 & 53.55 & 23.27 & 39.97 \\
    & OmniQuant~\cite{shao2023omniquant}      & 27.22 & 49.20 & 50.65 & 23.74 & 37.70 \\
    & AffineQuant~\cite{ma2024affinequant}    & 27.90 & 49.58 & 51.85 & 24.15 & 38.37 \\
    & \bf{\footnotesize \aespa}               & 33.36 & 55.64 & 58.31 & 23.12 & \bf{42.61} \\
    \midrule
    \multirowcell{7.5}{LLaMA-13B} & FP16 & 47.87 & 74.75 & 79.08 & 43.46 & 61.29 \\
    \cmidrule{2-7}
    & RTN & 28.16 & 27.15 & 26.09 & 25.53 & 26.73 \\
    & OPTQ~\cite{frantar2023optq}             & 27.22 & 25.76 & 25.67 & 25.05 & 25.93 \\
    & \zfold~\cite{jeon2023frustratingly}     & 32.68 & 58.08 & 57.89 & 26.44 & 43.77 \\
    & OmniQuant~\cite{shao2023omniquant}      & NaN & NaN & NaN & NaN & NaN \\
    & AffineQuant~\cite{ma2024affinequant}    & 32.17 & 56.36 & 60.29 & 25.22 & 43.51 \\
    & \bf{\footnotesize \aespa}               & 34.73 & 61.49 & 62.68 & 28.74 & \bf{46.91} \\
    \midrule
    \multirowcell{7.5}{LLaMA-30B} & FP16 & 52.90 & 78.96 & 82.63 & 54.66 & 67.29 \\
    \cmidrule{2-7}
    & RTN & 27.05 & 26.39 & 25.87 & 25.48 & 26.20 \\
    & OPTQ~\cite{frantar2023optq}             & 27.13 & 26.60 & 26.12 & 23.56 & 25.85 \\
    & \zfold~\cite{jeon2023frustratingly}     & 39.93 & 65.07 & 65.89 & 30.85 & 50.44 \\
    & OmniQuant~\cite{shao2023omniquant}      & 34.22 & 58.50 & 64.83 & 25.91 & 45.87 \\
    & AffineQuant~\cite{ma2024affinequant}    & NaN & NaN & NaN & NaN & NaN \\
    & \bf{\footnotesize \aespa}               & 41.13 & 67.00 & 67.90 & 35.67 & \bf{52.93} \\
    \midrule
    \multirowcell{7.5}{LLaMA2-7B} & FP16 & 46.16 & 74.49 & 75.99 & 41.87 & 59.63 \\
    \cmidrule{2-7}
    & RTN                                     & 28.33 & 26.01 & 25.88 & 23.02 & 25.81 \\
    & OPTQ~\cite{frantar2023optq}             & 26.37 & 26.09 & 25.11 & 25.10 & 25.67 \\
    & \zfold~\cite{jeon2023frustratingly}     & 26.62 & 42.68 & 44.71 & 22.88 & 34.22 \\
    & OmniQuant~\cite{shao2023omniquant}      & 25.00 & 38.80 & 42.97 & 23.03 & 32.45 \\
    & AffineQuant~\cite{ma2024affinequant}    & NaN & NaN & NaN & NaN & NaN \\
    & \bf{\footnotesize \aespa}               & 30.29 & 51.47 & 56.75 & 25.59 & \bf{41.03} \\
    \midrule
    \multirowcell{7.5}{LLaMA2-13B} & FP16 & 49.06 & 77.44 & 79.39 & 52.10 & 64.50 \\
    \cmidrule{2-7}
    & RTN                                     & 27.22 & 25.04 & 25.58 & 24.69 & 25.63 \\
    & OPTQ~\cite{frantar2023optq}             & 26.71 & 27.19 & 25.42 & 23.74 & 25.77 \\
    & \zfold~\cite{jeon2023frustratingly}     & 28.41 & 48.32 & 51.59 & 23.98 & 38.08 \\
    & OmniQuant~\cite{shao2023omniquant}      & 27.13 & 47.98 & 53.27 & 23.81 & 38.05 \\
    & AffineQuant~\cite{ma2024affinequant}    & 30.80 & 52.90 & 57.74 & 24.45 & 41.47 \\
    & \bf{\footnotesize \aespa}               & 31.91 & 55.18 & 55.49 & 29.97 & \bf{43.14} \\
    \bottomrule
    \end{tabular}
    \begin{tablenotes}
        \item[*] `NaN' means that loss diverges in the quantization process.
        \item[*] Results for high bit-widths are provided in~\cref{appendix:zero-shot performance} due to the page limitation.
    \end{tablenotes}
    \end{threeparttable}
    \label{tab:zero_shot-int2}

    \vspace{-.3cm}
    
\end{table*}

\textbf{Zero-shot task performance}
We evaluate the reasoning performance of quantized models using zero-shot tasks (\eg \ ARC~\cite{allenai:arc}, HellaSwag~\cite{zellers2019hellaswag}, and MMLU~\cite{hendrycks2020measuring}).
We note that the zero-shot setting was ensured in our experiments because we used excerpts from randomly crawled websites (not task-specific data) as a calibration dataset.
From the zero-shot results in \cref{tab:zero_shot-int2} and \cref{tab:zero-shot} (see \cref{appendix:zero-shot performance}), we observe that the proposed \aespa \ performs the best in almost all cases, and the performance gap between \aespa \ and the existing methods is large for 2-bit.

\textbf{Time cost}
We summarize the processing times of the different quantization algorithms in \cref{appendix:time-memory-costs}. 
We note that the processing time of \aespa \ includes the time required for pre-computations (lines 2-4 in \cref{algo:aespa}).
As expected, \aespa \ completes quantization much faster than \brecq. 
For example, while \brecq \ requires more than 10 GPU hours for OPT-1.3B, \aespa \ completes quantization in 1.24 hours, which demonstrates the effectiveness of the proposed pre-computation-based loss computation strategy.
Although other block-wise methods (OmniQuant/AffineQuant) perform quantization faster than \aespa \ for hyper-scale models, they suffer from unstable training processes or exhibit poor PPL performance (\eg \ PPL of OmniQuant is larger than $10^{3}$ for OPT-6.7B; see \cref{tab:comparison-with-block-wise-methods}). 
In addition, we observe that OPTQ performs quantization quickly, but its 2-bit performance collapses regardless of the model size (see \cref{tab:opt} in \cref{appendix:comparison with one-shot schemes}). 
Except for \aespa, \zfold \ is the only method that shows both reasonable performance and processing time.

\textbf{Discussion}
In real situations, when one needs to preserve the performance of the original model as much as possible, the proposed \aespa \ would be an intriguing solution. 
In particular, when deploying LLMs on resource-constrained platforms where up to 7B models are commonly employed (\eg \ mobile devices), \aespa \ would be a good fit.
Even when fast quantization of hyper-scale models is required, \aespa \ can be used with a slight modification. 
Specifically, in time-limited cases, one can skip the weight-rounding optimization (lines 5-8 in \cref{algo:aespa}) and simply determine the quantization parameters using the proposed Hessian that considers the cross-layer dependency (line~4 in \cref{algo:aespa}). 
In doing so, we can not only save the time required to optimize a weight-rounding mechanism, but also save the memory required to store pre-computed values ($\mathbb{E} [\mathbf{K}^{T} \mathbf{K}]$ and $\mathbb{E} [\mathbf{Q}^{T} \mathbf{Q}]$). 
Indeed, when performing only quantization parameter computation, we achieved a significant reduction in the processing time (see \cref{tab:aespa-time-no-rounding-optimization} in \cref{appendix:time-memory-costs}) while still exhibiting better performance than conventional methods (see \cref{tab:hessian_poc} in \cref{appendix:hessian_ablation}).

\section{Conclusion}  \label{sec:conclusion}

We proposed a next-level PTQ scheme for Transformers, called \aespa.
By targeting the attention-wise reconstruction while quantizing Transformers layer-wise, we could consider the cross-layer dependency within the attention module and complete the quantization much faster than the existing approach for block-wise reconstruction (\ie~\brecq).
Extensive experiments on language models have demonstrated the outstanding performance of \aespa.

\textbf{Limitations and future work} 
While we focused on the attention output, the output of the entire Transformer block (containing fully connected layers) can be used to consider the dependencies between more layers.
However, in this case, the objective functions would be more complicated than those in~\cref{eq:recon error_mid_Q} and~\cref{eq:recon error_mid_K} due to nonlinear activation functions (\eg \ SiLU for LLaMA models), normalization layers, and weights of larger dimensions.
Enhancing the quantization performance by developing an efficient form of the reconstruction error for the Transformer block would be an interesting future work.
Furthermore, while we focused on weight-only quantization, activations may need to be quantized to deploy AI models on integer-only arithmetic hardware (\eg \ NPU).
Extending the proposed \aespa \ for weight-activation quantization by integrating existing techniques to suppress activation outliers~\cite{xiao2023smoothquant, ashkboos2024quarot} is also an interesting research direction.
Finally, while we verified the performance of \aespa \ with LLMs, we believe that \aespa \ can also be used for the quantization of diffusion models.
To that end, we may need to incorporate some diffusion-specific quantization strategies to overcome output distribution discrepancies over different time steps (\eg \ grouping of time-steps with similar distributions~\cite{wang2024towards}, temporal feature preservation~\cite{huang2024tfmq}, and separate quantization for shortcuts in U-Net~\cite{li2023q}), which will be considered in our future studies.

\newpage

\bibliography{main}
\bibliographystyle{plain}

\newpage

\newpage
\appendix

\section*{Appendices}

\section{Pseudo-code for the proposed \aespa}  \label{appendix:pseudo-code}

In this appendix, we provide the pseudo-code for the proposed \aespa \ excluded in the main text due to the page limitation.

\begin{algorithm}[htb]
\begin{spacing}{1.1}
\footnotesize
\caption{Quantization} 
\label{algo:aespa}
\renewcommand\algorithmicrequire{\textbf{Input}:}
\renewcommand\algorithmicensure{\textbf{Output}:}
\renewcommand\algorithmicprocedure{\textbf{def} }
\begin{algorithmic}[1]
\Procedure{Quantization}{$\mW$,$\mX$}
    \State Approximate the Hessian $\mathbf{H}$ \Comment{See~\cref{tab:algo}}
    \State Estimate $\mathbb{E} [\mK^{T} \mK], \mathbb{E} [\mQ^{T} \mQ]$ for $\mW_{Q}, \mW_{K}$\Comment{\cref{tab:algo}}
    \State Set the step size $\mS$ s.t. $\min_{\mS} \trace \left ( \Delta \mW \mathbf{H} \Delta \mW^{T} \right )$
    \Repeat
        \State Compute the Loss $\gL$\Comment{\cref{tab:algo}}
        \State Optimize $\mS$ or $\mW_{int}$ w.r.t $\gL$ by certain algorithm
    \Until{converged}
    \State return $\mS$ and $\mW_{int}$\Comment{step size and integer weight}
\EndProcedure
\end{algorithmic}
\end{spacing}
\end{algorithm}
\begin{table}[htb]
    \renewcommand{\arraystretch}{1.0}
    \footnotesize
    \centering
    \vspace{-0.35cm}
    \caption{Row-wise Hessian $\mathbf{H}$ and quantization loss $\gL$ for each layer}
    \begin{tabular}{c c c}
        \toprule
        Layer                             &  $\mathbf{H}$   &  $\gL$  \\
        \toprule
        $\mW_Q$   & $\mathbb{E}\left [ \mX \mX^{T} \right ]$ 
                  &  $\trace \left ( \mathbb{E}\left[\mK^{T} \mK\right] \cdot \Delta \mW    \mathbf{H}  \Delta \mW^{T} \right )$ \\
        $\mW_K$   & $\mathbb{E}\left [ \mX \mX^{T} \right ]$ 
                  &  $\trace \left ( \mathbb{E}\left[\mQ^{T} \mQ\right] \cdot  \Delta \mW    \mathbf{H}  \Delta \mW^{T} \right )$ \\
        $\mW_V$   & $\mathbb{E}\left [\mX\mA^{T}\mA\mX^{T}\right ]$ 
                  &  $\trace \left ( \Delta \mW     \mathbf{H} \Delta \mW^{T} \right )$ \\
        Others    & $\mathbb{E}\left [ \mX \mX^{T} \right ]$ 
                  &  $\trace \left ( \Delta \mW   \mathbf{H}  \Delta \mW^{T} \right )$ \\
        \bottomrule
    \end{tabular}
    \label{tab:algo}
    \vspace{-0.25cm}
\end{table}

As mentioned, the proposed \aespa \ consists of two main steps; \aespa \ first determines the quantization parameters (\ie \  scale $s$ and zero-point $z$ in~\cref{eq:quantizer}) together with foldable parameters, as in~\cite{lin2023awq, jeon2023frustratingly, shao2023omniquant, ma2024affinequant} (see line 4 in \cref{algo:aespa}), and then optimizes an integer weight $\mW_{int}$ for each weight (see lines 5-8 in \cref{algo:aespa}).
We emphasize that each iteration for the integer weight optimization can be performed efficiently based on pre-computed values (\ie \  $\mathbb{E}[\mX \mX^{T}]$, $\mathbb{E}[\mX \mA^{T} \mA \mX^{T}]$, $\mathbb{E}[\mK^{T} \mK]$, and $\mathbb{E}[\mQ^{T} \mQ]$ in \cref{tab:algo}).
We also note that while we have used \zfold \ in computing the quantization parameters and used AdaRound in optimizing integer weights, our refined objectives in~\cref{eq:aespa_final recon error_V},~\cref{eq:quant goal_final_Q}, and~\cref{eq:quant goal_final_K} can be integrated into any layer-wise quantization scheme without effort because we only used the definition of the attention operation in our derivation.

\newpage

\section{Validity of the proposed separate quantization strategy}  \label{appendix: validity-of-qled}

\begin{table*}[htb]
    \renewcommand{\arraystretch}{1.2}
    \fontsize{7.5pt}{9pt}\selectfont
    \centering
    \caption{Performance (PPL $\downarrow$) of OPT-125M quantized with different strategies.}
    \vspace{-.2cm}
    \begin{tabular}{c c c c c c c c c c c}
        \toprule
        \multirow{2}{*}{Method} & \multirowcell{2}{Quantization \\ Granularity} & \multirowcell{2}{Reconstruction \\ Target} & \multicolumn{2}{c}{W2A16} & & \multicolumn{2}{c}{W3A16} & &\multicolumn{2}{c}{W4A16} \\
        \cline{4-5} \cline{7-8} \cline{10-11}
        &  &  & Wiki-2 & C4 & & Wiki-2 & C4 & & Wiki-2 & C4 \\
        \toprule
        AdaRound        & Layer-wise        & Layer Output          & 160.7 & 95.63 &   & 35.44 & 31.86 &   & 29.51 & 27.78 \\
        \brecq          & Block-wise        & Attention Output      & 60.38 & 47.85 &   & 33.25 & 29.74 &   & 28.86 & 27.43 \\
        \midrule
        \bf{Proposed}   & \bf{Layer-wise}   & \bf{Attention Output} & 69.23 & 51.92 &   & 32.89 & 29.75 &   & 28.98 & 27.42 \\
        \bottomrule
    \end{tabular}
    \label{tab:qled_poc}
\end{table*}

We conduct experiments to demonstrate the importance of targeting attention-wise reconstruction and the validity of the proposed separate quantization strategy.
In our experiments, we learn a weight-rounding policy using conventional AdaRound, but we set the loss function for each projection as the attention reconstruction error in~\cref{eq:attn recon error} (not the layer reconstruction error; see \cref{fig:qled}(c)).

\cref{tab:qled_poc} summarizes the quantization performance of AdaRound, \brecq, and our approach on the OPT-125M model.
As evident, our approach uniformly outperforms AdaRound for all bit-widths, although both methods quantize models layer-wise.
This is because we can consider the cross-layer dependency (\ie \  the relationship between the query, key, and value) by targeting attention-wise reconstruction, which differs from AdaRound wherein layers are considered independent.
Furthermore, once we target attention-wise reconstruction, separate layer-wise quantization does not incur severe performance degradation compared to the joint quantization method (\brecq).
Indeed, our approach causes only a marginal performance degradation for 2-bit and exhibits comparable performance for 3-bit and 4-bit.

One might wonder about the strategy of quantizing more than one layer jointly while maintaining remaining weights with full-precision, \eg \ simultaneous quantization of the query and key projections while fixing the value projection with full-precision.
To say the conclusion first, in this case, we cannot use the proposed pre-computation-based loss computation strategy (see \cref{subsec:algo}), resulting in a much longer quantization processing time.
Specifically, when quantizing $\mW_{Q}$ and $\mW_{K}$ simultaneously, the attention reconstruction error is expressed as
\begin{align*}
    \Delta \SA_{Q, K}
        &= \mathbb{E} \left [ \left \| \SA (\widehat{\mQ}, \widehat{\mK}, \mV) - \SA(\mQ, \mK, \mV) \right \|_{F}^{2} \right ]
        = \mathbb{E} \left [ \left \| \Delta \mA \mV \right \|_{F}^{2} \right ],
\end{align*}
where
\begin{align*}
    \Delta \mA
        &= \softmax \left ( \frac{\widehat{\mQ} \widehat{\mK}^{T}}{\sqrt{d}} \right ) - \softmax \left ( \frac{\mQ \mK^{T}}{\sqrt{d}} \right ).
\end{align*}
Then, by taking similar steps as in \cref{subsec:refined objectives} (\ie \  approximating $\Delta \mA$ with its first-order Taylor series and constructing an upper bound of $\Delta \SA_{Q, K}$), we can obtain the following objective:
\begin{align*}
    \lefteqn{\min_{\Delta \mW_{Q}, \Delta \mW_{K}} \mathbb{E} \left [ \left \| \widehat{\mQ} \widehat{\mK}^{T} - \mQ \mK^{T} \right \|_{F}^{2} \right ]} \nonumber \\
        &~~~= \min_{\Delta \mW_{Q}, \Delta \mW_{K}} \mathbb{E} \left [ \left \| \Delta \mQ \mK^{T} + \mQ \Delta \mK^{T} + \Delta \mQ \Delta \mK^{T} \right \|_{F}^{2} \right ] \nonumber \\
        &~~~= \min_{\Delta \mW_{Q}, \Delta \mW_{K}} \mathbb{E} \left [ \left \| \mX^{T} \Delta \mW_{Q}^{T} \mK^{T} + \mQ \Delta \mW_{K} \mX + \mX^{T} \Delta \mW_{Q}^{T} \Delta \mW_{K} \mX \right \|_{F}^{2} \right ].
\end{align*}
Obviously, the objective becomes much more complex than the proposed ones in~\cref{eq:quant goal_initial_Q} and~\cref{eq:quant goal_initial_K}, and it would be difficult to simplify and accelerate the loss computation by exploiting pre-computed values as in \aespa. 
In fact, without the proposed pre-computation-based loss computation, the simultaneous quantization of $\mW_{Q}$ and $\mW_{K}$ requires 3.5 hours for the quantization of OPT-125M, which is about 44 times longer than the proposed \aespa \ and even 1.9 times longer than \brecq.

\newpage

\section{Refined quantization objective~\cref{eq:quant goal_initial_K} for the key projection}  \label{appendix: proof_K}

When quantizing the key projection $\mW_{K}$, we fix the query and value projections with full-precision.
In this case, the attention reconstruction error $\Delta \SA_{K}$ can be expressed as
\begin{align*}
    \Delta \SA_{K}
        &= \mathbb{E} \left [ \left \| \SA ( \mQ, \widehat{\mK}, \mV ) - \SA ( \mQ, \mK, \mV ) \right \|_{F}^{2} \right ]
        = \mathbb{E} \left [ \left \| \Delta \mA \mV \right \|_{F}^{2} \right ],
\end{align*}
where
\begin{align*}
    \Delta \mA
        &= \softmax \left ( \frac{\mQ \widehat{\mK}^{T}}{\sqrt{d}} \right ) - \softmax \left ( \frac{\mQ \mK^{T}}{\sqrt{d}} \right ).
\end{align*}
To avoid the computational overhead of repetitive softmax operation, we approximate $\Delta \mA$ with its first-order Taylor series, which leads to
\begin{align}  \label{eq:recon error_mid_K}
    \Delta \SA_{K}
        &\approx \frac{1}{d} \mathbb{E} \left [ \left \| \mQ \Delta \mK^{T} \mathbf{J}_{\softmax}^{T} \mV \right \|_{F}^{2}\right ]
        = \frac{1}{d} \mathbb{E} \left [ \left \| \mQ \Delta \mW_{K} \mX \mathbf{J}_{\softmax}^{T} \mV \right \|_{F}^{2}\right ].
\end{align}
Furthermore, to reduce the huge memory cost required to store the Jacobian $\mathbf{J}_{\softmax}$ having $L^{3}$ elements (see \cref{footnote:jacobian_shape}), we establish an upper bound of~\cref{eq:recon error_mid_K} and then use it as a surrogate of $\Delta \SA_{K}$.
Specifically, we separate the term $\| \mQ \Delta \mW_{K} \mX \mathbf{J}_{\softmax}^{T} \mV \|_{F}^{2}$ into two components as follows:
\begin{align*}
    \left \| \mQ \Delta \mW_{K} \mX \mathbf{J}_{\softmax}^{T} \mV \right \|_{F}^{2}
        &\le \left \| \mQ \Delta \mW_{K} \mX \right \|_{F}^{2} \cdot \left \| \mathbf{J}_{\softmax}^{T} \mV \right \|_{F}^{2}.
\end{align*}
Noting that the term $\left \| \mathbf{J}_{\softmax}^{T} \mV \right \|_{F}^{2}$ is not affected by the quantization of $\mW_{K}$ and thus fixed in the quantization process, we minimize $\| \mQ \Delta \mW_{K} \mX \|_{F}^{2}$ to enforce $\Delta \SA_{K}$ to be small, which leads to the proposed objective in~\cref{eq:quant goal_initial_K}.

\newpage

\section{Effectiveness of the proposed Hessian in~\cref{eq:aespa_Hessian_V}}  \label{appendix:hessian_ablation}

We recall from \cref{subsec:algo} that the proposed quantization objective for the value projection is
\begin{align*}
    \trace \left ( \Delta \mW_{V} \mathbb{E} \left [ \mX \mA^{T} \mA \mX^{T} \right ] \Delta \mW_{V}^{T} \right ),
\end{align*}
which implies that the Hessian $\mathbf{H}_{V}$ for each row of $\mW_{V}$ is
\begin{align*}
    \mathbf{H}_{V} = 2\mathbb{E} [ \mX \mA^{T} \mA \mX^{T} ].
\end{align*}
We note that the proposed Hessian $\mathbf{H}_{V}$ differs from 
\begin{align*}
    \mathbf{H} = 2\mathbb{E} [\mX \mX^{T}],
\end{align*} 
which has been commonly used as an approximated Hessian in existing methods~\cite{obq, frantar2023optq, chee2023quip, jeon2023frustratingly}.
The key reason for the difference is that we consider the dependency between the query, key, and value projections by targeting attention-wise reconstruction, whereas the previous methods assumed independence.

To observe the effect of considering the cross-layer dependency, we use different Hessians (\ie \  $\mathbf{H}_{V}$ and $\mathbf{H}$) when quantizing language models via \zfold \ and then compare the performance of the quantized models.
As \cref{tab:hessian_poc} shows, the quantization performance is much better when the proposed Hessian $\mathbf{H}_{V}$ is used, which demonstrates the importance of considering the cross-layer dependency.

\begin{table*}[htb]
    \renewcommand{\arraystretch}{1.0}
    \fontsize{7.8pt}{9.36pt}\selectfont
    \centering
    \caption{Quantization performance (PPL $\downarrow$) of \zfold \ under different Hessians.}
    \vspace{-.1cm}
    \begin{subtable}{\textwidth}
        \centering
        \caption{WikiText-2}
        \vspace{-.2cm}
        \begin{tabular}{l c c c c c c c c c c}
        \toprule
        \multirow{2}{*}{Hessian} & \multirow{2}{*}{Precision} & \multicolumn{5}{c}{OPT} & & \multicolumn{3}{c}{LLaMA} \\
        \cline{3-7} \cline{9-11}
        & & \multicolumn{1}{c}{125M} & \multicolumn{1}{c}{350M} & \multicolumn{1}{c}{1.3B} & \multicolumn{1}{c}{2.7B} & \multicolumn{1}{c}{6.7B} & & \multicolumn{1}{c}{7B} & \multicolumn{1}{c}{13B} & \multicolumn{1}{c}{30B} \\
        \toprule
        $\mathbb{E} [\mX \mX^{T}]$~\cite{obq, frantar2023optq, chee2023quip, jeon2023frustratingly} & \multirow{2}{*}{INT3} & 39.59 & 25.97 & 16.10 & 13.54 & 11.65 & & 6.756 & 5.708 & 4.931 \\
        \bf{$\mathbb{E} [\mX \mA^{T} \mA \mX^{T}]$ (ours)}                                          &                        & \bf{35.05} & \bf{24.81} & 16.25 & \bf{13.40} & \bf{11.43} & & \bf{6.529} & \bf{5.669} & \bf{4.693} \\
        \midrule
        $\mathbb{E} [\mX \mX^{T}]$~\cite{obq, frantar2023optq, chee2023quip, jeon2023frustratingly} & \multirow{2}{*}{INT2} & 190.1 & 102.5 & 33.97 & 27.10 & 18.07 & & 14.93 & \bf{13.03} & 9.250 \\
        \bf{$\mathbb{E} [\mX \mA^{T} \mA \mX^{T}]$ (ours)}                                          &                        & \bf{146.4} & \bf{68.30} & \bf{31.43} & \bf{25.17} & \bf{17.92} & & \bf{14.20} & 13.15 & \bf{8.138} \\
        \bottomrule
        \end{tabular}
    \end{subtable}

    \vspace{.2cm}
    
    \begin{subtable}{\textwidth}
        \centering
        \caption{PTB}
        \vspace{-.2cm}
        \begin{tabular}{l c c c c c c c c c c}
        \toprule
        \multirow{2}{*}{Hessian} & \multirow{2}{*}{Precision} & \multicolumn{5}{c}{OPT} & & \multicolumn{3}{c}{LLaMA} \\
        \cline{3-7} \cline{9-11}
        & & \multicolumn{1}{c}{125M} & \multicolumn{1}{c}{350M} & \multicolumn{1}{c}{1.3B} & \multicolumn{1}{c}{2.7B} & \multicolumn{1}{c}{6.7B} & & \multicolumn{1}{c}{7B} & \multicolumn{1}{c}{13B} & \multicolumn{1}{c}{30B} \\
        \toprule
        $\mathbb{E} [\mX \mX^{T}]$~\cite{obq, frantar2023optq, chee2023quip, jeon2023frustratingly} & \multirow{2}{*}{INT3} & 53.08 & 39.23 & 22.73 & 20.18 & 16.64 & & 11.73 & \bf{10.09} & 8.979 \\
        \bf{$\mathbb{E} [\mX \mA^{T} \mA \mX^{T}]$ (ours)}                                          &                        & \bf{49.88} & \bf{37.62} & \bf{22.66} & \bf{19.78} & \bf{16.55} & & \bf{11.39} & 10.48 & \bf{8.657} \\
        \midrule
        $\mathbb{E} [\mX \mX^{T}]$~\cite{obq, frantar2023optq, chee2023quip, jeon2023frustratingly} & \multirow{2}{*}{INT2} & 331.6 & 130.7 & 53.80 & 46.08 & 26.79 & & 26.87 & 19.37 & 15.23 \\
        \bf{$\mathbb{E} [\mX \mA^{T} \mA \mX^{T}]$ (ours)}                                          &                        & \bf{212.8} & \bf{100.1} & \bf{53.64} & \bf{42.93} & \bf{26.09} & & \bf{24.88} & \bf{18.01} & \bf{12.99} \\
        \bottomrule
        \end{tabular}
    \end{subtable}
    
    \vspace{.2cm}

    \begin{subtable}{\textwidth}
        \centering
        \caption{C4}
        \vspace{-.2cm}
        \begin{tabular}{l c c c c c c c c c c}
        \toprule
        \multirow{2}{*}{Hessian} & \multirow{2}{*}{Precision} & \multicolumn{5}{c}{OPT} & & \multicolumn{3}{c}{LLaMA} \\
        \cline{3-7} \cline{9-11}
        & & \multicolumn{1}{c}{125M} & \multicolumn{1}{c}{350M} & \multicolumn{1}{c}{1.3B} & \multicolumn{1}{c}{2.7B} & \multicolumn{1}{c}{6.7B} & & \multicolumn{1}{c}{7B} & \multicolumn{1}{c}{13B} & \multicolumn{1}{c}{30B} \\
        \toprule
        $\mathbb{E} [\mX \mX^{T}]$~\cite{obq, frantar2023optq, chee2023quip, jeon2023frustratingly} & \multirow{2}{*}{INT3} & 33.67 & 26.45 & 17.33 & 15.50 & 13.28 & & 8.719 & 7.554 & 6.912 \\
        \bf{$\mathbb{E} [\mX \mA^{T} \mA \mX^{T}]$ (ours)}                                          &                        & \bf{31.27} & \bf{25.51} & \bf{17.27} & \bf{15.42} & \bf{13.22} & & \bf{8.313} & \bf{7.437} & \bf{6.638} \\
        \midrule
        $\mathbb{E} [\mX \mX^{T}]$~\cite{obq, frantar2023optq, chee2023quip, jeon2023frustratingly} & \multirow{2}{*}{INT2} & 125.3 & 71.37 & 31.67 & 25.99 & 19.79 & & 16.88 & 14.61 & 11.90 \\
        \bf{$\mathbb{E} [\mX \mA^{T} \mA \mX^{T}]$ (ours)}                                          &                        & \bf{112.6} & \bf{56.48} & \bf{30.06} & \bf{25.34} & \bf{19.32} & & \bf{16.87} & \bf{13.46} & \bf{10.32} \\
        \bottomrule
        \end{tabular}
    \end{subtable}
    
    \label{tab:hessian_poc}
\end{table*}

\newpage

\section{Proof of~\cref{eq:quant goal_mid_Q} and~\cref{eq:quant goal_final_Q}} \label{appendix: proof_Q}

Note that $\mathbb{E} \left [ \left \| \mK \Delta \mW_{Q} \mX \right \|_{F}^{2} \right ] =\mathbb{E} \left [ \left \| \vect \left ( \mK \Delta \mW_{Q} \mX \right ) \right \|_{2}^{2} \right ]$, where $\vect(\cdot)$ denotes the vectorization operation.
Then, by exploiting the following properties of Kronecker product
\begin{align}
    \vect \left ( \mA \mB \mC \right ) &= \left ( \mC^{T} \otimes \mA \right ) \vect (\mB), \nonumber \\
    \left ( \mA \otimes \mB \right )^{T} &= \mA^{T} \otimes \mB^{T}, \nonumber \\
    \left ( \mA \otimes \mB \right ) \left ( \mC \otimes \mD \right ) &= \mA \mC \otimes \mB \mD, \nonumber 
\end{align}
we have
\begin{align}
    \mathbb{E} \left [ \left \| \mK \Delta \mW_{Q} \mX \right \|_{F}^{2} \right ]
        &= \mathbb{E} \left [ \left \| \left ( \mX^{T} \otimes \mK \right ) \Delta \vw_{Q} \right \|_{2}^{2} \right ] \nonumber \\
        &= \mathbb{E} \left [ \Delta \vw_{Q}^{T} \left ( \mX^{T} \otimes \mK \right )^{T} \left ( \mX^{T} \otimes \mK \right ) \Delta \vw_{Q} \right ] \nonumber \\
        &= \mathbb{E} \left [ \Delta \vw_{Q}^{T} \left ( \mX \otimes \mK^{T} \right ) \left ( \mX^{T} \otimes \mK \right ) \Delta \vw_{Q} \right ] \nonumber \\
        &= \mathbb{E} \left [ \Delta \vw_{Q}^{T} \left ( \mX \mX^{T} \otimes \mK^{T} \mK \right ) \Delta \vw_{Q} \right ] \label{eq:appendix 1} \\
        &= \Delta \vw_{Q}^{T} \cdot \mathbb{E} \left [ \mX \mX^{T} \otimes \mK^{T} \mK \right ] \cdot \Delta \vw_{Q}, \nonumber
\end{align}
which is the desired result in~\cref{eq:quant goal_mid_Q}.

We now prove~\cref{eq:quant goal_final_Q}.
By combining~\cref{eq:quant goal_mid_Q} and~\cref{eq:kronecker approximation}, we have
\begin{align*}
    \mathbb{E} \left [ \left \| \mK \Delta \mW_{Q} \mX \right \|_{F}^{2} \right ]
        &\approx \Delta \vw_{Q}^{T} \cdot \left ( \mathbb{E} \left [ \mX \mX^{T} \right ] \otimes \mathbb{E} \left [ \mK^{T} \mK \right ] \right ) \cdot \Delta \vw_{Q}. 
\end{align*}
Note that since $\mathbb{E} [ \mX \mX^{T} ]$ and $\mathbb{E} [ \mK^{T} \mK ]$ are symmetric, there exist $\mG_{X}$ and $\mG_{K}$ such that 
\begin{align*}
    \mathbb{E} [ \mX \mX^{T} ] = \mG_{X} \mG_{X}^{T},~\mathbb{E} [ \mK^{T} \mK ] = \mG_{K}^{T} \mG_{K}.
\end{align*}
Then, by following the steps used to derive~\cref{eq:appendix 1} in the reverse order, we have
\begin{align}
    \mathbb{E} \left [ \left \| \mK \Delta \mW_{Q} \mX \right \|_{F}^{2} \right ]
        &= \Delta \vw_{Q}^{T} \left ( \mG_{X} \mG_{X}^{T} \otimes \mG_{K}^{T} \mG_{K} \right ) \Delta \vw_{Q} \nonumber \\
        &= \left \| \mG_{K} \Delta \mW_{Q} \mG_{X} \right \|_{F}^{2} \nonumber \\
        &= \trace \left ( \mG_{K} \Delta \mW_{Q} \mG_{X} \mG_{X}^{T} \Delta \mW_{Q}^{T} \mG_{K}^{T} \right ) \nonumber \\
        &= \trace \left ( \mG_{K}^{T} \mG_{K} \cdot \Delta \mW_{Q} \cdot \mG_{X} \mG_{X}^{T} \cdot \Delta \mW_{Q}^{T} \right ) \nonumber \\
        &= \trace \left ( \mathbb{E} \left [ \mK^{T} \mK \right ] \Delta \mW_{Q} \mathbb{E} \left [ \mX \mX^{T} \right ] \Delta \mW_{Q}^{T} \right ), \nonumber
\end{align}
which completes the proof of~\cref{eq:quant goal_final_Q}.

\newpage

\section{Integration of proposed loss functions into existing PTQ schemes}  \label{appendix: integration_adaround}

We recall that we only utilized the definition of the attention operation when developing the proposed loss functions for the attention output reconstruction.
Therefore, our loss functions can be integrated into any PTQ schemes based on layer-wise reconstruction and used to enhance their performance.
In this section, we describe how to combine our loss functions with existing quantization schemes by taking AdaRound as an example.

In short, AdaRound learns a weight-rounding mechanism by solving the following optimization problem~\cite{nagel2020up}:
\begin{align}  \label{eq:adaround}
    \arg \min_{\mB}~\left \| \mW \mX - \widetilde{\mW} \mX \right \|_{F}^{2} + \lambda \sum_{i, j} \left ( 1 - \left | 2 h (\mB_{i, j}) - 1 \right |^{\beta} \right ),
\end{align}
where $\mB$ is the continuous variable to be learned, $h$ is the rectified sigmoid function, and $\widetilde{\mW}$ is the soft-quantized weights defined as 
\begin{align*}
    \widetilde{\mW}
        &= s \cdot \text{clamp} \left ( \left \lfloor \frac{\mW}{s} \right \rfloor + h(\mB), n, p \right ).
\end{align*}
One can see that the loss function of AdaRound consists of two components, layer-wise reconstruction error and weight-rounding loss.

To consider the cross-layer dependency between $\mW_{Q}$, $\mW_{K}$, and $\mW_{V}$ in the learning process, we integrate the proposed loss functions developed for the attention output reconstruction into~\cref{eq:adaround}.
In other words, we replace the layer-wise reconstruction error in~\cref{eq:adaround} with our loss functions in~\cref{eq:aespa_final recon error_V},~\cref{eq:quant goal_final_Q}, and~\cref{eq:quant goal_final_K}.
For example, when learning the rounding policy for the query projection matrix $\mW_{Q}$, the objective of the proposed \aespa \ is expressed as
\begin{align}  \label{eq:aespa_adaround}
    \arg \min_{\mB_{Q}}~\trace \hspace{-.5mm} \left ( \mathbb{E} \hspace{-.5mm} \left [ \mK^{T} \mK \right ] \hspace{-.5mm} \Delta \mW_{Q} \mathbb{E} \left [ \mX \mX^{T} \right ] \hspace{-.5mm} \Delta \mW_{Q}^{T} \right ) + \lambda \sum_{i, j} \left ( 1 - \left | 2 h (\mB_{Q, i, j}) - 1 \right |^{\beta} \right ),
\end{align}
where $\Delta \mW_{Q} = \mW_{Q} - \widetilde{\mW}_{Q}$.

\newpage

\section{Complexity analysis for conventional block-wise quantization schemes}  \label{appendix:complexity_conventional}

Recall from~\cref{eq:attn recon error} that conventional block-wise quantization schemes require to compute $\SA (\widehat{\mQ}, \widehat{\mK}, \widehat{\mV})$ in each iteration. 
This means that for each input sequence, one needs to perform 
\begin{itemize} [leftmargin=1.5em]
    \item forward pass for $\widehat{\mQ}$, $\widehat{\mK}$, and $\widehat{\mV}$: $3d_{h}L(2d-1)$ flops
    
    \item matrix multiplications for computing $\widehat{\mQ}\widehat{\mK}^{T}$ and $\widehat{\mA} \widehat{\mV}$: $4d_{h}L^{2} -d_{h}L -L^{2}$ flops

    \item softmax operation with additional scaling (\ie \  $\softmax ( \frac{\widehat{\mQ}\widehat{\mK}^{T}}{\sqrt{d_{h}}})$): $3L^{2} + d_{h}L - L$ flops

    \item final computation of reconstruction error: $3d_{h}L - 1$ flops
\end{itemize}
If $B$ input sequences are used in each quantization iteration, then the total number of flops required in conventional methods is
\begin{align*}
    \gC_{exist} 
        = B(6d_{h}dL + 4d_{h}L^{2} +2L^{2} - L - 1) 
        = \gO(Bd_{h}L \cdot \max \{ d, L \}).
\end{align*}

\paragraph{Comparison of $\gC_{\aespa}$ and $\gC_{exist}$}
We now compare the complexities of \aespa~and conventional block-wise quantization methods in terms of the number of flops.
\cref{tab:complexity} summarizes the computational costs required to quantize different sizes of OPT models.
For conventional methods, we report the cost of using four sequences in each iteration ($B=4$).
We observe that the computational cost of \aespa~is considerably lower than that of conventional methods.
In particular, for small-scale models (\eg \ OPT-125M, OPT-350M, and OPT-1.3B), \aespa~performs ten times fewer number of flops.
One can notice that the gap between $\gC_{\aespa}$ and $\gC_{exist}$ decreases as the model size increases.
This is because the hidden size $d$ exceeds the sequence length $L$ (which is fixed for all models) as the model size increases.
Nevertheless, \aespa~still incurs a lower computational cost, and the gap increases if conventional methods use larger batch sizes.

\begin{table}[htb]
    \renewcommand{\arraystretch}{1.2}
    \footnotesize
    \centering
    \caption{Cost of \aespa~and conventional methods (GFLOPS)}
    \begin{tabular}{c c c c c c c}
        \toprule
                    &   125M    &   350M    & 1.3B  & 2.7B  & 6.7B  & 13B  \\
        \toprule
        $\mathcal{C}_{exist}$    & 6.7     & 7.5     & 11      & 15      & 34    & 41 \\
        $\mathcal{C}_{\aespa}$  & 0.24    & 0.42    & 1.6     & 3.2     & 13    & 20 \\
        \bottomrule
    \end{tabular}
    \vspace{-0.25cm}
    \label{tab:complexity}
\end{table}

\newpage

\section{Comparison with block-wise PTQ schemes}  \label{appendix:comparison with block-wise schemes_PTB}

We provide experimental results excluded from the main text due to page limitations.

\begin{table*}[htb]
    \renewcommand{\arraystretch}{1.0}
    \fontsize{7.5pt}{9.0pt}\selectfont
    \centering
    \caption{Performance (PPL $\downarrow$) of the proposed \aespa~and conventional block-wise PTQ methods.}
    \vspace{-.1cm}
    \begin{subtable}{\textwidth}
    \centering
    \caption{INT4 performance on WikiText-2 and C4}
    \vspace{-.2cm}
    \begin{tabular}{c l c c c c c c c c c c c}
    \toprule
    \multirow{2}{*}{Dataset} & \multirow{2}{*}{Method} & \multicolumn{4}{c}{OPT} & & \multicolumn{3}{c}{LLaMA} & & \multicolumn{2}{c}{LLaMA2}\\
    \cline{3-6} \cline{8-10} \cline{12-13}
    & & 125M & 1.3B & 2.7B & 6.7B & & 7B & 13B & 30B & & 7B & 13B \\
    \toprule
    \multirowcell{5.5}{WikiText-2}
    & FP16 & 27.65 & 14.63 & 12.47 & 10.86 &   & 5.677 & 5.091 & 4.101 & & 5.472 & 4.884 \\
    \cmidrule{2-13}
    & \brecq~\cite{li2021brecq}            & \bf{28.86} & 14.83 & 12.71 & OOM & & OOM & OOM & OOM & & OOM & OOM \\
    & OmniQuant~\cite{shao2023omniquant}   & 30.42 & 15.15 & 12.89 & 11.20 & & 5.907 & 5.256 & 4.263 & & 5.850 & 5.064 \\
    & AffineQuant~\cite{ma2024affinequant} & 29.81 & 15.09 & 12.72 & 11.12 & & 5.905 & 5.256 & 4.269 & & 5.782 & 5.062 \\
    & \bf{\footnotesize \aespa}            & 28.87 & \bf{14.81} & \bf{12.36} & \bf{10.95} & & \bf{5.890} & \bf{5.226} & \bf{4.254} & & \bf{5.684} & \bf{5.031} \\
    \midrule
    \multirowcell{5.5}{C4}
    & FP16 & 26.56 & 16.07 & 14.34 & 12.71 &   & 7.344  & 6.798 & 6.131 & & 7.264 & 6.727 \\
    \cmidrule{2-13}
    & \brecq~\cite{li2021brecq}            & 27.43 & 16.42 & 14.61 & OOM & & OOM & OOM & OOM & & OOM & OOM \\
    & OmniQuant~\cite{shao2023omniquant}   & 28.41 & 16.68 & 14.83 & 12.99 & & 7.656 & 6.976 & 6.269 & & 7.686 & 6.956 \\
    & AffineQuant~\cite{ma2024affinequant} & 28.04 & 16.58 & 14.74 & 12.92 & & 7.654 & 6.974 & 6.270 & & 7.644 & 6.927 \\
    & \bf{\footnotesize \aespa}            & \bf{27.24} & \bf{16.31} & \bf{14.55} & \bf{12.82} & & \bf{7.633} & \bf{6.945} & \bf{6.256} & & \bf{7.508} & \bf{6.891} \\
    \bottomrule
    \end{tabular}
    \end{subtable}

    \vspace{.2cm}
    
    \begin{subtable}{\textwidth}
    \centering
    \caption{Performance on PTB}
    \vspace{-.2cm}
    \begin{threeparttable}
    \begin{tabular}{c l c c c c c c c c}
    \toprule
    \multirow{2}{*}{Precision} & \multirow{2}{*}{Method} & \multicolumn{4}{c}{OPT} & & \multicolumn{3}{c}{LLaMA} \\
    \cline{3-6} \cline{8-10}
    & & 125M & 1.3B & 2.7B & 6.7B & & 7B & 13B & 30B \\
    \toprule
    FP16 & Baseline & 38.99 & 20.29 & 17.97 & 15.77 &   & 10.12 & 9.081 & 8.159 \\
    \midrule
    \multirow{4}{*}{INT4}
    & \brecq~\cite{li2021brecq}            & 41.04 & 20.97 & 18.41 & OOM & & OOM & OOM & OOM \\
    & OmniQuant~\cite{shao2023omniquant}   & 42.34 & 21.32 & 18.70 & 16.04 & & 10.57 & 9.330 & 8.354 \\
    & AffineQuant~\cite{ma2024affinequant} & 42.99 & 21.26 & 18.49 & 16.02 & & 10.53 & 9.325 & 8.355  \\
    & \bf{\footnotesize \aespa}            & \bf{40.50} & \bf{20.78} & \bf{18.30} & \bf{15.84} & & \bf{10.43} & \bf{9.277} & \bf{8.283} \\
    \midrule
    \multirow{4}{*}{INT3}
    & \brecq~\cite{li2021brecq}            & 46.93 & 23.41 & 19.82 & OOM & & OOM & OOM & OOM \\
    & OmniQuant~\cite{shao2023omniquant}   & 56.88 & 25.11 & 22.59 & 18.33 & & 11.98 & 10.24 & 9.065 \\
    & AffineQuant~\cite{ma2024affinequant} & 51.47 & 24.38 & 21.03 & 17.40 & & 11.92 & 10.24 & 8.998 \\
    & \bf{\footnotesize \aespa}            & \bf{44.96} & \bf{22.35} & \bf{19.48} & \bf{16.28} & & \bf{11.45} & \bf{9.818} & \bf{8.684} \\
    \midrule
    \multirow{4}{*}{INT2}
    & \brecq~\cite{li2021brecq}            & \bf{90.22} & 344.9 & 282.0 & OOM & & OOM & OOM & OOM \\
    & OmniQuant~\cite{shao2023omniquant}   & NaN & 377.9 & 2.0e3 & 7.7e3 & & 33.51 & NaN & 17.38 \\
    & AffineQuant~\cite{ma2024affinequant} & 177.8 & 75.25 & 47.07 & 37.90 & & 29.33 & 18.58 & NaN \\
    & \bf{\footnotesize \aespa}            & 99.12 & \bf{37.19} & \bf{32.57} & \bf{22.80} & & \bf{19.83} & \bf{15.65} & \bf{12.98} \\
    \bottomrule
    \end{tabular}
    \begin{tablenotes}
        \item[*] `NaN' means that loss diverges in the quantization process.
        \item[*] `OOM' means that out-of-memory issues occur when quantizing models with a single A100 GPU.
    \end{tablenotes}
    \end{threeparttable}
    \end{subtable}
    
    \label{tab:comparison-with-block-wise-methods-4bit}
    
\end{table*}

\newpage

\section{Comparison with layer-wise PTQ schemes}  \label{appendix:comparison with one-shot schemes}

We provide experimental results excluded from the main text due to page limitations.

\subsection{Results on OPT models}

\begin{table*}[htb!]
    \renewcommand{\arraystretch}{1.0}
    \small
    \centering
    \caption{Performance (PPL $\downarrow$) of \aespa~and existing layer-wise PTQ methods on OPT models.}
    \vspace{-.1cm}
    \begin{subtable}{\textwidth}
        \centering
        \caption{WikiText-2}
        \vspace{-.2cm}
        \begin{tabular}{c l c c c c c c c}
        \toprule
        Precision & Method & 125M & 350M & 1.3B & 2.7B & 6.7B & 13B & 30B \\
        \toprule
        FP16 & Baseline & 27.65 & 22.00 & 14.63 & 12.47 & 10.86 & 10.13 & 9.56 \\
        \midrule
        \multirow{4}{*}{INT4}
        & RTN                     & 37.28 & 25.94 & 48.20 & 16.92 & 12.10 & 11.32 & 10.98 \\
        & OPTQ                    & 32.49 & 23.68 & 15.50 & 12.85 & 11.12 & 10.33 & 9.670 \\
        & \zfold                  & 31.03 & 23.08 & 15.00 & 12.47 & 11.01 & 10.21 & 9.537  \\
        & \bf{\normalsize \aespa} & \bf{28.87} & \bf{22.55} & \bf{14.81} & \bf{12.36} & \bf{10.95} & \bf{10.18} &  \bf{9.511} \\
        \midrule
        \multirow{4}{*}{INT3}
        & RTN                     & 1.3e3	& 64.57	& 1.3e4	& 1.6e4	& 5.8e3	& 3.4e3	& 1.6e3 \\
        & OPTQ                    & 52.95	& 33.29	& 20.36	& 16.94	& 13.01	& 11.65	& 10.44 \\
        & \zfold                  & 39.59 & 25.97 & 16.10 & 13.54 & 11.65 & 10.62 & 9.902 \\
        & \bf{\normalsize \aespa} & \bf{32.71} & \bf{24.45} & \bf{15.79} & \bf{13.14} & \bf{11.23} & \bf{10.52}    &   \bf{9.760}   \\
        \midrule
        \multirow{4}{*}{INT2}
        & RTN                     & 5.5e3 & 2.8e4 & 1.1e5 & 9.5e3 & 2.8e4 & 1.9e5 & 1.7e5 \\
        & OPTQ                    & 4.1e3	& 1.1e4	& 8.3e3	& 9.3e3	& 2.0e3	& 539.8	& 56.63 \\
        & \zfold                  & 190.1 & 102.5 & 33.97 & 27.10 & 18.07 & 33.48 & 13.48 \\
        & \bf{\normalsize \aespa} & \bf{71.18} & \bf{54.89} & \bf{24.26} & \bf{22.22} & \bf{15.71} & \bf{15.27}    & \bf{11.91}  \\
        \bottomrule
        \end{tabular}
        \label{tab:opt_wikitext2}
    \end{subtable}

    \vspace{.2cm}
    
    \begin{subtable}{\textwidth}
        \centering
        \caption{PTB}
        \vspace{-.2cm}
        \begin{tabular}{c l c c c c c c c}
        \toprule
        Precision & Method & 125M & 350M & 1.3B & 2.7B & 6.7B & 13B & 30B \\
        \toprule
        FP16 & Baseline & 38.99 & 31.08 & 20.29 & 17.97 & 15.77 & 14.52 & 14.04 \\
        \midrule
        \multirow{4}{*}{INT4}
        & RTN                     & 53.88 & 36.79 & 75.37 & 32.41 & 18.86 & 16.41 & 15.44 \\
        & OPTQ                    & 46.54 & 33.27 & 21.74 & 19.04 & 16.42 & 14.88 & 14.21 \\
        & \zfold                  & 44.17 & 33.51 & 20.96 & 18.45 & 15.98 & 14.65 & 14.11  \\
        & \bf{\normalsize \aespa} & \bf{40.50} & \bf{32.17} & \bf{20.78} & \bf{18.30} & \bf{15.84} & \bf{14.65}   & \bf{14.09}  \\
        \midrule
        \multirow{4}{*}{INT3}
        & RTN                     & 1.4e3	& 87.21	& 1.5e4	& 1.4e4	& 5.3e3	& 2.2e3	& 1.5e3 \\
        & OPTQ                    & 74.07	& 46.10	& 29.76	& 25.06	& 19.22	& 16.42	& 15.08 \\
        & \zfold                  & 53.08 & 39.23 & 22.73 & 20.18 & 16.64 & 15.23 & 14.60 \\
        & \bf{\normalsize \aespa} & \bf{44.96} & \bf{36.15} & \bf{22.35} & \bf{19.48} & \bf{16.28} & \bf{15.06}    & \bf{14.43}    \\
        \midrule
        \multirow{4}{*}{INT2}
        & RTN                     & 4.3e3 & 2.8e4 & 1.1e4 & 6.8e3 & 1.8e4 & 1.2e5 & 1.7e5 \\
        & OPTQ                    & 3.5e3	& 1.2e4	& 6.6e3	& 8.0e3	& 2.5e3	& 458.4	& 83.81 \\
        & \zfold                  & 331.6 & 130.7 & 53.80 & 46.08 & 26.79 & 79.69 & 20.39 \\
        & \bf{\normalsize \aespa} & \bf{99.12} & \bf{79.86} & \bf{37.19} & \bf{32.57} & \bf{22.80} & \bf{23.93}    & \bf{17.51}   \\
        \bottomrule
        \end{tabular}
        \label{tab:opt_ptb}
    \end{subtable}

    \vspace{.2cm}
    
    \begin{subtable}{\textwidth}
        \centering
        \caption{C4}
        \vspace{-.2cm}
        \begin{tabular}{c l c c c c c c c}
        \toprule
        Precision & Method & 125M & 350M & 1.3B & 2.7B & 6.7B & 13B & 30B \\
        \toprule
        FP16 & Baseline & 26.56 & 22.59 & 16.07 & 14.34 & 12.71 & 12.06 & 11.44 \\
        \midrule
        \multirow{4}{*}{INT4}
        & RTN                     & 33.88	& 26.21	& 27.50	& 18.83	& 14.37	& 13.32	& 13.55 \\
        & OPTQ                    & 29.64	& 24.15	& 16.75	& 14.86	& 13.00	& 12.24	& 11.56 \\
        & \zfold                  & 28.92 & 23.71 & 16.38 & 14.60 & 12.85 & 12.14 & 11.49  \\
        & \bf{\normalsize \aespa} & \bf{27.24} & \bf{23.15} & \bf{16.31} & \bf{14.55} & \bf{12.82} & \bf{12.13}  & \bf{11.47}  \\
        \midrule
        \multirow{4}{*}{INT3}
        & RTN                     & 834.4	& 55.15	& 6.6e3	& 1.2e4	& 5.0e3	& 2.8e3	& 1.8e3 \\
        & OPTQ                    & 42.88	& 30.60	& 20.53	& 17.66	& 14.61	& 13.19	& 12.15 \\
        & \zfold                  & 33.67 & 26.45 & 17.33 & 15.50 & 13.28 & 12.45 & 11.73 \\
        & \bf{\normalsize \aespa} & \bf{29.51} & \bf{24.96} & \bf{17.10} & \bf{15.27} & \bf{13.15} & \bf{12.39}    &   \bf{11.68}  \\
        \midrule
        \multirow{4}{*}{INT2}
        & RTN                     & 3.7e3 & 1.6e4 & 7.7e3 & 7.7e3 & 1.4e4 & 9.7e4 & 5.8e4 \\
        & OPTQ                    & 2.1e3	& 4.4e3	& 3.0e3	& 3.7e3	& 290.9	& 157.7	& 29.40 \\
        & \zfold                  & 125.3 & 71.37 & 31.67 & 25.98 & 19.79 & 47.10 & 14.51 \\
        & \bf{\normalsize \aespa} & \bf{56.88} & \bf{46.36} & \bf{23.54} & \bf{22.53} & \bf{17.28} & \bf{16.30}    & \bf{13.32}   \\
        \bottomrule
        \end{tabular}
        \label{tab:opt_c4}
    \end{subtable}
    
    \label{tab:opt}
    
\end{table*}

\newpage

\subsection{Results on BLOOM models}

\begin{table*}[hbt!]
    \renewcommand{\arraystretch}{1.0}
    \fontsize{7.5pt}{9.0pt}\selectfont
    \centering
    \caption{Performance (PPL $\downarrow$) of \aespa~and existing layer-wise PTQ methods on BLOOM models.}
    \vspace{-.1cm}
    \begin{subtable}{\textwidth}
        \centering
        \caption{INT4 performance on WikiText-2 and C4}
        \vspace{-.2cm}
        \begin{tabular}{c l c c c c c c c c c c c}
        \toprule
        \multirowcell{2}{Precision} & \multirowcell{2}{Method} & \multicolumn{5}{c}{WikiText-2} & & \multicolumn{5}{c}{C4}\\
        \cline{3-7} \cline{9-13}
        & & 560M & 1.1B & 1.7B & 3B & 7.1B & & 560M & 1.1B & 1.7B & 3B & 7.1B \\
        \toprule
        FP16 & Baseline & 22.42 & 17.69 & 15.39 & 13.48 & 11.37 & & 26.60 & 22.05 & 19.49 & 17.49 & 15.20 \\
        \midrule
        \multirowcell{4}{INT4}
        & RTN                     & 25.82	& 19.98	& 16.96	& 14.75	& 12.09 & & 29.80 & 24.42 & 21.24 & 18.75 & 16.05 \\
        & OPTQ                    & 23.83	& 18.74	& 16.16	& 14.01	& 11.72 & & 27.74 & 23.05 & 20.26 & 18.00 & 15.54 \\
        & \zfold                  & 23.60 & 18.44 & 15.87 & 13.90 & 11.59 & & 27.36 & 22.66 & 20.00 & 17.87 & 15.42 \\
        & \bf{\footnotesize\aespa}  & \bf{23.21} & \bf{18.28} & \bf{15.76} & \bf{13.81} & \bf{11.56} & & \bf{27.20} & \bf{22.49} & \bf{19.86} & \bf{17.76} & \bf{15.38} \\
        \bottomrule
        \end{tabular}
    \end{subtable}
    
    \vspace{.2cm}
    
    \begin{subtable}{\textwidth}
        \centering
        \subcaption{Performance on PTB}
        \vspace{-.2cm}
        \begin{tabular}{c l c c c c c}
        \toprule
        Precision & Method & 560M & 1.1B & 1.7B & 3B & 7.1B \\
        \toprule
        FP16 & Baseline & 43.69 & 57.96 & 30.00 & 25.34 & 20.83 \\
        \midrule
        \multirow{4}{*}{INT4}
        & RTN                     & 50.96	& 66.79	& 33.52	& 27.65	& 22.40 \\
        & OPTQ                    & 46.83	& 62.99	& 31.63	& 26.72	& 21.52 \\
        & \zfold                  & 45.77 & 61.33 & 31.26 & 26.27 & 21.35  \\
        & \bf{\footnotesize\aespa}  & \bf{44.73}    & \bf{60.41}    & \bf{31.05}    & \bf{26.01}    & \bf{21.17}    \\
        \midrule
        \multirow{4}{*}{INT3}
        & RTN                     & 124.8	& 184.0	& 105.5	& 66.24	& 34.94 \\
        & OPTQ                    & 64.43	& 82.91	& 40.27	& 33.13	& 25.94 \\
        & \zfold                  & 53.01 & 69.93 & 35.12 & 28.41 & 22.83 \\
        & \bf{\footnotesize\aespa}  & \bf{48.87}    & \bf{67.01}    & \bf{33.06}    & \bf{27.61}   & \bf{22.03}    \\
        \midrule
        \multirow{4}{*}{INT2}
        & RTN                     & 7.4e5	& 1.1e6	& 2.5e5	& 1.2e5	& 2.2e5 \\
        & OPTQ                    & 4.1e3	& 2.4e3	& 1.4e3	& 1.4e3	& 428.4 \\
        & \zfold                  & 194.9 & 174.9 & 74.03 & 69.49 & 38.50 \\
        & \bf{\footnotesize\aespa}  & \bf{91.14}    & \bf{120.7}    & \bf{57.48}    & \bf{46.40}    & \bf{31.28}    \\
        \bottomrule
        \end{tabular}
        \label{tab:bloom_ptb}
    \end{subtable}
    
    \label{tab:bloom}
    
\end{table*}

\newpage

\subsection{Results on LLaMA models}

\begin{table*}[hbt!]
    \renewcommand{\arraystretch}{1.0}
    \centering
    \caption{Performance (PPL $\downarrow$) of \aespa~and existing layer-wise PTQ methods on LLaMA models.}
    \fontsize{7.5pt}{9.0pt}\selectfont
    \vspace{-.2cm}
    \begin{tabular}{c l c c c c c c c c c c c c}
    \toprule
    \multirow{2}{*}{Precision} & \multirow{2}{*}{Method} &  & \multicolumn{3}{c}{WikiText-2} &  & \multicolumn{3}{c}{PTB} &     & \multicolumn{3}{c}{C4} \\
    \cline{4-6} \cline{8-10} \cline{12-14}
    & & & 7B & 13B & 30B & & 7B & 13B & 30B & & 7B & 13B & 30B \\
    \toprule
    FP16 & Baseline &   & 5.677 & 5.091 & 4.101 &  & 10.12 & 9.081 & 8.159  &  & 7.344  & 6.798 & 6.131    \\
    \midrule
    \multirow{4}{*}{INT4}
    & RTN                         & & 6.291 & 5.525 & 4.536 &   & 11.25 & 9.775 & 8.653 &   & 8.121 & 7.232 & 6.537 \\
    & OPTQ                        & & 6.167 & 5.365 & 4.452 &   & 11.51 & 9.526 & 8.426 &   & 7.792 & 7.082 & 6.399 \\
    & \zfold                      & & 6.069 & 5.278 & 4.325 & & 11.45 & 9.335 & 8.410 & & 7.797 & 6.984 & 6.318 \\
    & \bf{\footnotesize \aespa}   & & \bf{5.890} & \bf{5.226} & \bf{4.254} & & \bf{10.43} & \bf{9.277} & \bf{8.283} & & \bf{7.633} & \bf{6.945} & \bf{6.256} \\
    \midrule
    \multirow{4}{*}{INT3}
    & RTN                         & & 25.61 & 11.78 & 14.87 &   & 98.89 & 28.94 & 28.79 &   & 30.86 & 14.46 & 30.04 \\
    & OPTQ                        & & 8.290 & 6.729 & 5.705 &   & 16.11 & 11.91 & 9.964 &   & 10.51 & 8.832 & 7.977 \\
    & \zfold                      & & 6.756 & 5.708 & 4.931 & & 11.73 & 10.09 & 8.979 & & 8.719 & 7.554 & 6.912 \\
    & \bf{\footnotesize \aespa}   & & \bf{6.579} & \bf{5.611} & \bf{4.688} & & \bf{11.45} & \bf{9.818} & \bf{8.684} & & \bf{8.465} & \bf{7.399} & \bf{6.634} \\
    
    \midrule
    \multirow{4}{*}{INT2}
    & RTN                         & & 1.1e5 & 5.7e4 & 2.7e4 &   & 9.9e4 & 8.1e4 & 3.3e4 &   & 1.1e5 & 5.9e4 & 2.8e4 \\
    & OPTQ                        & & 1.0e4 & 3.7e3 & 1.5e3 &   & 1.1e4 & 8.5e3 & 1.0e3 &   & 872.7 & 809.7 & 304.4 \\
    & \zfold                      & & 14.93 & 13.03 & 9.250 & & 26.87 & 19.37 & 15.23 & & 16.88 & 14.61 & 11.90 \\
    & \bf{\footnotesize \aespa}   & & \bf{11.94} & \bf{10.30} & \bf{7.845} & & \bf{19.83} & \bf{15.65} & \bf{12.98} & & \bf{13.63} & \bf{11.46} & \bf{10.35} \\
    \bottomrule
    \end{tabular}

    \label{tab:llama}
    
\end{table*}

\subsection{Results on LLaMA2 models}

\begin{table}[htb]
    \renewcommand{\arraystretch}{1.0}
    \fontsize{7.5pt}{9.0pt}\selectfont
    \centering
    \caption{Performance (PPL $\downarrow$) of \aespa~and existing layer-wise PTQ methods on LLaMA2 models.}
    \begin{tabular}{c l c c c c c c c}
    \toprule
    \multirow{2}{*}{Precision} & \multirow{2}{*}{Method} &  & \multicolumn{2}{c}{WikiText-2} &  & \multicolumn{2}{c}{C4} \\
    \cline{4-5} \cline{7-8}
    & & & 7B & 13B & & 7B & 13B \\
    \toprule
    FP16 & Baseline & 
    & 5.472 & 4.884 & & 7.264 & 6.727 \\
    \midrule
    \multirowcell{4}{INT4}
    & RTN       & & 6.116 & 5.205 & & 8.165 & 7.142 \\
    & OPTQ      & & 6.060 & 5.158 & & 7.838 & 7.030 \\
    & \zfold    & & 5.815 & 5.099 & & 7.602 & 6.996 \\
    & \textbf{\footnotesize \aespa} & & \bf{5.684} & \bf{5.031} & & \bf{7.508} & \bf{6.891} \\
    \midrule
    \multirowcell{4}{INT3}
    & RTN       & & 542.0 & 10.69 & & 527.2 & 13.87 \\
    & OPTQ      & & 8.664 & 6.554 & & 11.24 & 8.761 \\
    & \zfold    & & 6.606 & 5.710 & & 8.666 & 7.692 \\
    & \textbf{\footnotesize \aespa} & & \bf{6.241} & \bf{5.462} & & \bf{8.225} & \bf{7.392} \\
    \midrule
    \multirowcell{4}{INT2}
    & RTN       & & 1.8e4 & 5.1e4 & & 2.8e4 & 5.3e4 \\
    & OPTQ      & & 7.5e3 & 2.1e3 & & 1.7e3 & 560.7 \\
    & \zfold    & & 20.79 & 15.56 & & 21.98 & 16.90 \\
    & \textbf{\footnotesize \aespa} & & \bf{13.99} & \bf{12.14} & & \bf{14.36} & \bf{13.59} \\
    \bottomrule                                                       
    \end{tabular}

    \label{tab:llama2}
    
\end{table}

\newpage

\section{Results for zero-shot tasks}  \label{appendix:zero-shot performance}

We provide INT3 zero-shot performance results that are excluded from the main text due to page limitations.

\begin{table*}[htb]
    \renewcommand{\arraystretch}{1.0}
    \fontsize{7.5pt}{9.0pt}\selectfont
    \centering
    \caption{INT3 zero-shot performance (accuracy $\uparrow$) of \aespa \ and existing methods.}
    \vspace{-.2cm}
    \begin{tabular}{c l c c c c c}
    \toprule
    Model & Method & ARC-c & ARC-e & HellaSwag & MMLU & Average \\
    \toprule
    \multirowcell{7.5}{LLaMA-7B}
    & FP16 & 44.62 & 72.85 & 76.18 & 32.19 & 56.46 \\
    \cmidrule{2-7}
    & RTN & 27.47 & 45.45 & 45.46 & 24.94 & 35.83 \\
    & OPTQ~\cite{frantar2023optq}             & 36.95 & 62.63 & 68.33 & 25.51 & 48.36 \\
    & \zfold~\cite{jeon2023frustratingly}     & 41.21 & 66.92 & 72.50 & 28.90 & 52.38 \\
    & OmniQuant~\cite{shao2023omniquant}      & 38.99 & 67.30 & 70.31 & 29.33 & 51.48 \\
    & AffineQuant~\cite{ma2024affinequant}    & 39.25 & 65.61 & 70.56 & 29.68 & 51.28 \\
    & \bf{\small \aespa}                      & 40.87 & 69.15 & 71.54 & 30.57 & \bf{53.03} \\
    \midrule
    \multirowcell{7.5}{LLaMA-13B}
    & FP16 & 47.87 & 74.75 & 79.08 & 43.46 & 61.29 \\
    \cmidrule{2-7}
    & RTN & 36.09 & 56.23 & 62.03 & 26.20 & 45.14 \\
    & OPTQ~\cite{frantar2023optq}             & 43.00 & 67.89 & 72.45 & 28.62 & 52.99 \\
    & \zfold~\cite{jeon2023frustratingly}     & 44.88 & 71.00 & 75.66 & 36.88 & 57.11 \\
    & OmniQuant~\cite{shao2023omniquant}      & 44.03 & 69.70 & 75.15 & 35.89 & 56.19 \\
    & AffineQuant~\cite{ma2024affinequant}    & 43.60 & 70.24 & 75.10 & 32.67 & 55.40 \\
    & \bf{\small \aespa}                      & 45.82 & 71.80 & 75.87 & 38.63 & \bf{58.03} \\
    \midrule
    \multirowcell{7.5}{LLaMA-30B}
    & FP16 & 52.90 & 78.96 & 82.63 & 54.66 & 67.29 \\
    \cmidrule{2-7}
    & RTN & 27.90 & 43.64 & 31.42 & 23.34 & 31.58 \\
    & OPTQ~\cite{frantar2023optq}             & 45.31 & 71.55 & 77.17 & 42.01 & 59.01 \\
    & \zfold~\cite{jeon2023frustratingly}     & 50.34 & 75.84 & 79.69 & 51.00 & 64.22 \\
    & OmniQuant~\cite{shao2023omniquant}      & 49.49 & 76.52 & 79.76 & 50.68 & 64.11 \\
    & AffineQuant~\cite{ma2024affinequant}    & 49.66 & 77.10 & 79.49 & 50.37 & 64.16 \\
    & \bf{\small \aespa}                      & 50.34 & 77.53 & 79.79 & 50.55 & \bf{64.55} \\
    \midrule
    \multirowcell{7.5}{LLaMA2-7B} 
    & FP16                                    & 46.16 & 74.49 & 75.99 & 41.87 & 59.63 \\
    \cmidrule{2-7}
    & RTN                                     & 25.94 & 35.48 & 35.39 & 23.14 & 29.99 \\
    & OPTQ~\cite{frantar2023optq}             & 37.46 & 63.01 & 64.85 & 28.79 & 48.53 \\
    & \zfold~\cite{jeon2023frustratingly}     & 40.10 & 64.65 & 69.92 & 33.69 & 52.09 \\
    & OmniQuant~\cite{shao2023omniquant}      & 40.36 & 67.30 & 71.00 & 31.26 & 52.48 \\
    & AffineQuant~\cite{ma2024affinequant}    & 40.78 & 67.21 & 70.75 & 30.93 & 52.42 \\
    & \bf{\footnotesize \aespa}               & 41.38 & 69.11 & 71.78 & 38.18 & \bf{55.11} \\
    \midrule
    \multirowcell{7.5}{LLaMA2-13B} 
    & FP16                                    & 49.06 & 77.44 & 79.39 & 52.10 & 64.50 \\
    \cmidrule{2-7}
    & RTN                                     & 34.56 & 55.98 & 59.44 & 25.45 & 43.86 \\
    & OPTQ~\cite{frantar2023optq}             & 43.09 & 70.45 & 72.02 & 39.37 & 56.23 \\
    & \zfold~\cite{jeon2023frustratingly}     & 46.42 & 72.77 & 74.79 & 47.91 & 60.47 \\
    & OmniQuant~\cite{shao2023omniquant}      & 45.65 & 74.33 & 74.77 & 43.92 & 59.67 \\
    & AffineQuant~\cite{ma2024affinequant}    & 47.18 & 75.42 & 75.28 & 45.61 & 60.87 \\
    & \bf{\footnotesize \aespa}               & 46.84 & 75.25 & 75.78 & 47.09 & \bf{61.24} \\
    \bottomrule
    \end{tabular}
    
    \label{tab:zero-shot}
    
\end{table*}

\newpage

\section{Time and memory cost comparison}  \label{appendix:time-memory-costs}

\begin{table*}[htb]
    \renewcommand{\arraystretch}{1.0}
    \fontsize{7.5pt}{9.0pt}\selectfont
    \centering
    \caption{Time and memory cost of \aespa \ and existing methods}

    \vspace{-.1cm}
    
    \begin{subtable}{\textwidth}
        \centering
        \caption{INT2 quantization processing time}
        \vspace{-.2cm}
        \begin{tabular}{c l c c c c c c c c}
        \toprule
        \multirow{2}{*}{Target} & \multirow{2}{*}{Method} & \multicolumn{4}{c}{OPT} & & \multicolumn{3}{c}{LLaMA} \\
        \cline{3-6} \cline{8-10}
        & & 125M & 1.3B &  2.7B & 6.7B & & 7B & 13B & 30B \\
        \toprule
        \multirowcell{2}{layer-wise \\ reconstruction}
        & OPTQ~\cite{frantar2023optq}           & 0.66 min  & 0.08 hr   & 0.14 hr   & 0.29 hr   &   & 0.25 hr   & 0.45 hr   & 1.08 hr   \\
        & \zfold~\cite{jeon2023frustratingly}   & 1.09 min & 0.27 hr & 0.61 hr & 2.58 hr &    & 1.13 hr & 2.48 hr & 10.51 hr \\
        \midrule
        \multirowcell{4}{attention-wise \\ reconstruction}
        & \brecq~\cite{li2021brecq}             & 108.2 min & 10.71 hr  & 19.15 hr  & OOM & & OOM & OOM & OOM \\
        & OmniQuant~\cite{shao2023omniquant}    & 16.20 min & 1.02 hr   & 1.63 hr   & 2.93 hr   &   & 2.37 hr  & 4.20 hr    & 9.84 hr   \\
        & AffineQuant~\cite{ma2024affinequant}  & 28.33 min & 2.57 hr   & 4.60 hr  & 9.85 hr  &   & 10.09 hr  & 18.76 hr  & 47.84 hr  \\
        & \bf{\small \aespa}                    & 4.78 min & 1.24 hr & 2.83 hr & 10.24 hr &   & 6.84 hr   & 15.89 hr   & 53.69 hr   \\
        \bottomrule
        \end{tabular}
    \end{subtable}

    \vspace{.2cm}
    
    \begin{subtable}{\textwidth}
        \centering
        \caption{Memory cost (GB)}
        \vspace{-.2cm}
        \begin{threeparttable}
        \begin{tabular}{c l c c c c c c c c}
        \toprule
        \multirow{2}{*}{Target} & \multirow{2}{*}{Method} & \multicolumn{4}{c}{OPT} & & \multicolumn{3}{c}{LLaMA} \\
        \cline{3-6} \cline{8-10}
        & & 125M & 1.3B &  2.7B & 6.7B & & 7B & 13B & 30B \\
        \toprule
        \multirowcell{2}{layer-wise \\ reconstruction}
        & OPTQ~\cite{frantar2023optq}           & 1.39 & 4.49 & 6.43 & 13.07 & & 8.76 & 12.34 & 18.59 \\
        & \zfold~\cite{jeon2023frustratingly}   & 1.39 & 4.49 & 6.43 & 13.07 & & 8.76 & 12.34 & 18.59 \\
        \midrule
        \multirowcell{4}{attention-wise \\ reconstruction}
        & \brecq~\cite{li2021brecq}             & 3.39  & 16.60 & 27.79 & OOM & & OOM & OOM & OOM  \\
        & OmniQuant~\cite{shao2023omniquant}    & 1.94  & 5.87  & 7.09  & 11.68 &   & 12.61 & 17.02 & 24.53 \\
        & AffineQuant~\cite{ma2024affinequant}  & 3.47  & 9.96   & 12.25 & 20.08 &   & 24.28 & 27.10 & 38.59 \\
        & \bf{\small \aespa}                    & 1.68  & 5.47  & 6.84  & 12.26  &   & 21.69 & 29.27 & 43.00 \\
        \bottomrule
        \end{tabular}
        \begin{tablenotes}
            \item[*] `OOM' means that out-of-memory issues occur when quantizing models with a single NVIDIA A100 GPU.
        \end{tablenotes}
        \end{threeparttable}
    \end{subtable}
    
    \label{tab:time-comparison}
    
\end{table*}

\cref{tab:time-comparison} summarizes the processing time and memory cost of different quantization algorithms.
We note that the processing time of the proposed \aespa \ includes the time required for pre-computations (lines 2-4 in \cref{algo:aespa}).

As expected, \aespa \ completes quantization much faster than \brecq. 
For example, while \brecq \ requires more than 10 hours to quantize OPT-1.3B, \aespa \ completes quantization in 1.24 hours, which demonstrates the effectiveness of the proposed objectives and pre-computation-based loss computation strategy.
Although other block-wise PTQ methods (OmniQuant/AffineQuant) perform quantization faster than \aespa \ for hyper-scale models, they suffer from unstable training process or exhibit poor PPL performance (\eg \ PPL of OmniQuant is larger than $10^{3}$ for OPT-6.7B; see \cref{tab:comparison-with-block-wise-methods}). 
We also observe that OPTQ performs quantization very fast, but its PPL performance collapses completely regardless of the model size (see \cref{tab:opt}). 
Except \aespa, \zfold \ is the only method that shows both reasonable performance and processing time.

In real situations, when one needs to preserve the performance of the original model as much as possible, the proposed \aespa \ would be an intriguing solution. 
In particular, when deploying LLMs on resource-constrained platforms where up to 7B models are commonly employed (\eg \ mobile devices), \aespa \ would be a good fit.
Even when fast quantization of hyper-scale models is needed, \aespa \ can be used with a slight modification. 
Specifically, in time-limited cases, one can skip weight-rounding optimization (lines 5-8 in \cref{algo:aespa}) and simply perform the quantization parameter computation (line 4 in \cref{algo:aespa}) using the proposed Hessian that considers the cross-layer dependency (see~\cref{eq:aespa_Hessian_V}). 
In doing so, we can not only save the time required to perform weight-rounding learning, but also save the memory required to store pre-computed values ($\mathbb{E} [\mathbf{K}^{T} \mathbf{K}]$ and $\mathbb{E} [\mathbf{Q}^{T} \mathbf{Q}]$). 
Indeed, when performing only quantization parameter computation, we achieved a significant reduction in the processing time (see \cref{tab:aespa-time-no-rounding-optimization} below) while still exhibiting better performance than conventional methods (see \cref{tab:hessian_poc} in \cref{appendix:hessian_ablation}).

\begin{table*}[htb]
    \renewcommand{\arraystretch}{1.0}
    \fontsize{7.5pt}{9.0pt}\selectfont
    \centering
    \caption{INT2 quantization processing time of \aespa \ without weight-rounding optimization}
    \vspace{-.1cm}
    \begin{tabular}{c c c c c c c c}
    \toprule
    \multicolumn{4}{c}{OPT} & & \multicolumn{3}{c}{LLaMA} \\
    \cline{1-4} \cline{6-8}
    \multicolumn{1}{c}{125M} & \multicolumn{1}{c}{1.3B} & \multicolumn{1}{c}{2.7B} & \multicolumn{1}{c}{6.7B} & & \multicolumn{1}{c}{7B} & \multicolumn{1}{c}{13B} & \multicolumn{1}{c}{30B} \\
    \toprule
    1.29 min & 0.35 hr & 0.74 hr & 2.92 hr & & 1.47 hr & 3.26 hr & 12.50 hr \\
    \bottomrule
    \end{tabular}
    
    \label{tab:aespa-time-no-rounding-optimization}
\end{table*}

\newpage

\section{Experimental results for different calibration datasets}  \label{appendix:wiki2}

One might wonder why the PPL performances of OmniQuant summarized in \cref{tab:comparison-with-block-wise-methods} are much worse than those reported in the original paper~\cite{shao2023omniquant}; INT2 PPL performances of quantized LLaMA models are 18.18, NaN, and 10.15 for WikiText-2 in \cref{tab:comparison-with-block-wise-methods}, which are worse than the values (15.47, 13.21, and 8.71) reported in~\cite{shao2023omniquant}.
This is because we used a different calibration dataset for quantization.
Specifically, we used C4 when constructing a calibration dataset, while~\cite{shao2023omniquant} used WikiText-2.

Additionally, we evaluate the performance of the proposed \aespa \ using WikiText-2 as a calibration dataset.
From~\cref{tab:wiki2}, we observe that when calibration data are sampled from WikiText-2, our results for OmniQuant are comparable with those reported in the original paper~\cite{shao2023omniquant}.
While it has been reported that the performance variance of OmniQuant across different calibration datasets is low for INT3 and INT4 (see~\cite[Table A10]{shao2023omniquant}), such low variance does not hold for INT2.
Furthermore, we observe that the proposed \aespa \ outperforms OmniQuant regardless of the type of the calibration dataset.

\begin{table}[htb]
    \renewcommand{\arraystretch}{1.0}
    \small
    \centering
    \caption{INT2 performances (PPL $\downarrow$) of \aespa \ and OmniQuant for different calibration datasets}
    \begin{threeparttable}
    \begin{tabular}{c c c c c c}
        \toprule
            \multirowcell{2}{Calibration \\ Dataset} & \multirowcell{2}{Method} & & \multicolumn{3}{c}{LLaMA} \\
            \cline{4-6}
            & & & 7B & 13B & 30B \\
        \toprule
            \multirowcell{2.5}{C4}
            & OmniQuant &
            & 18.18 & NaN & 10.15 \\
            \cmidrule{2-6}
            & \textbf{\aespa} &
            & \bf{11.94} & \bf{10.30} & \bf{7.845} \\
        \midrule
            \multirowcell{2.5}{WikiText-2}
            & OmniQuant &
            & 15.59 & 13.76 & 9.230 \\
            \cmidrule{2-6}
            & \textbf{\aespa} &
            & \textbf{8.818} & \textbf{7.423} & \textbf{6.232} \\
        \bottomrule                                
    \end{tabular}
    \begin{tablenotes}
        \item[*] Test dataset: WikiText-2
    \end{tablenotes}
    \end{threeparttable}
    \label{tab:wiki2}
\end{table}

\section{Quantization performance of \aespa \ for high bit-widths}  \label{appendix:INT4_INT6}

While previous results demonstrate that the proposed \aespa \ is very competitive for low-bit quantization (\eg \ INT2 and INT3), one might wonder whether \aespa \ can preserve the performance of the original full-precision model at high bit-widths.
We thus evaluate INT4 and INT6 quantization performances of \aespa \ with LLaMA models.
From~\cref{tab:INT6}, we observe that \aespa \ almost preserves the performance of the original full-precision model for the INT6 quantization.
Even for the INT4 quantization, the performance degradation is very marginal (\eg \ less than 1\% degradation for 13B and 30B models).

\begin{table}[htb]
    \renewcommand{\arraystretch}{1.0}
    \fontsize{8.33pt}{10.0pt}\selectfont
    \centering
    \caption{INT4 and INT6 quantization performances of the proposed \aespa \ (calibration data: C4)}
    \vspace{-.01cm}
    \begin{tabular}{c c c c c c c c c c c}
        \toprule
            \multirowcell{2}{Model} & \multirowcell{2}{Precision} & & \multicolumn{2}{c}{Perplexity ($\downarrow$)} & & \multicolumn{5}{c}{Zero-shot Accuracy ($\uparrow$)} \\
            \cline{4-5} \cline{7-11}
            & & & Wiki-2 & C4 & & ARC-c & ARC-e & HellaSwag & MMLU & Average \\
        \toprule
            \multirowcell{4}{LLaMA-7B} 
            & FP16 & 
            & 5.677 & 7.344 &  
            & 44.62 & 72.85 & 76.18 & 32.19 & 56.46 \\
            \cmidrule{2-11}
            & INT4 & 
            & 5.896 & 7.602 & 
            & 43.77 & 71.51 & 74.90 & 31.33 & 55.38 \\
            & INT6 & 
            & 5.694 & 7.360 & 
            & 44.62 & 72.35 & 75.96 & 32.27 & 56.30 \\
        \midrule
            \multirowcell{4}{LLaMA-13B} 
            & FP16 & 
            & 5.091 & 6.798 & 
            & 47.87 & 74.75 & 79.08 & 43.46 & 61.29 \\
            \cmidrule{2-11}
            & INT4 & 
            & 5.232 & 6.938 & 
            & 47.53 & 73.74 & 78.35 & 43.49 & 60.78 \\
            & INT6 & 
            & 5.096 & 6.809 & 
            & 48.04 & 74.96 & 78.98 & 43.24 & 61.31 \\
        \midrule
            \multirowcell{4}{LLaMA-30B} 
            & FP16 & 
            & 4.101 & 6.131 &
            & 52.90 & 78.96 & 82.63 & 54.66 & 67.29 \\
            \cmidrule{2-11}
            & INT4 & 
            & 4.260 & 6.254 &
            & 52.99 & 78.16 & 82.28 & 53.62 & 66.76 \\
            & INT6 & 
            & 4.110 & 6.139 &
            & 53.07 & 78.96 & 82.60 & 54.61 & 67.31 \\
        \bottomrule                                                        
    \end{tabular}

    \label{tab:INT6}
\end{table}

\newpage

\section{Experimental results for different seeds}  \label{appendix:seed}

We recall that when constructing a calibration dataset, we randomly draw 128 sequences from the C4 dataset~\cite{c4}.
By changing the seed for the sampling, different calibration datasets can be constructed, which leads to different quantization results.
In this appendix, we report the corresponding results and overall statistics.

\begin{table*}[htb!]
    \renewcommand{\arraystretch}{1.0}
    \fontsize{7.8pt}{9.36pt}\selectfont
    \centering
    \caption{Quantization performance (PPL $\downarrow$) of \aespa~on OPT models for different seeds.}
    \vspace{-.1cm}
    \begin{subtable}{\textwidth}
        \centering
        \caption{WikiText-2}
        \vspace{-.2cm}
        \begin{tabular}{c c c c c c c}
        \toprule
        Precision & Seed & 125M & 350M & 1.3B & 2.7B & 6.7B \\
        \toprule
        \multirowcell{3}{INT4} 
        & 0     & 28.87 & 22.55 & 14.81 & 12.36 & 10.95 \\
        & 10    & 28.60 & 22.55 & 14.91 & 12.31 & 10.83 \\
        & 100   & 28.75 & 22.85 & 14.94 & 12.35 & 10.90 \\ 
        \midrule
        \multirowcell{3}{INT3} 
        & 0     & 32.71 & 24.45 & 15.79 & 13.14 & 11.23 \\
        & 10    & 32.95 & 24.57 & 16.10 & 13.21 & 11.11 \\
        & 100   & 33.38 & 24.45 & 15.70 & 13.27 & 11.24 \\
        \midrule
        \multirowcell{3}{INT2} 
        & 0     & 71.18 & 54.89 & 24.26 & 22.22 & 15.71 \\
        & 10    & 74.41 & 50.84 & 24.38 & 22.36 & 15.06 \\
        & 100   & 77.03 & 53.12 & 25.93 & 22.39 & 15.66 \\
        \bottomrule
        \end{tabular}
    \end{subtable}

    \vspace{.2cm}

    \begin{subtable}{\textwidth}
        \centering
        \caption{PTB}
        \vspace{-.2cm}
        \begin{tabular}{c c c c c c c}
        \toprule
        Precision & Seed & 125M & 350M & 1.3B & 2.7B & 6.7B \\
        \toprule
        \multirowcell{3}{INT4} 
        & 0     & 40.50 & 32.17 & 20.78 & 18.30 & 15.84 \\
        & 10    & 40.62 & 32.33 & 20.56 & 18.21 & 15.91 \\
        & 100   & 40.11 & 32.60 & 20.55 & 18.20 & 15.86 \\
        \midrule
        \multirowcell{3}{INT3} 
        & 0     & 44.96 & 36.15 & 22.35 & 19.48 & 16.28 \\
        & 10    & 46.26 & 36.19 & 22.06 & 19.46 & 16.32 \\
        & 100   & 47.54 & 35.61 & 22.10 & 19.66 & 16.39 \\
        \midrule
        \multirowcell{3}{INT2} 
        & 0     & 99.12 & 79.86 & 37.19 & 32.57 & 22.80 \\
        & 10    & 110.0 & 73.98 & 35.94 & 32.25 & 21.51 \\
        & 100   & 106.0 & 79.09 & 37.33 & 31.90 & 21.86 \\
        \bottomrule
        \end{tabular}
    \end{subtable}

    \vspace{.2cm}

    \begin{subtable}{\textwidth}
        \centering
        \caption{C4}
        \vspace{-.2cm}
        \begin{tabular}{c c c c c c c}
        \toprule
        Precision & Seed & 125M & 350M & 1.3B & 2.7B & 6.7B \\
        \toprule
        \multirowcell{3}{INT4} 
        & 0     & 27.24 & 23.15 & 16.31 & 14.55 & 12.82 \\
        & 10    & 27.23 & 23.13 & 16.32 & 14.54 & 12.81 \\
        & 100   & 27.29 & 23.15 & 16.34 & 14.54 & 12.81 \\
        \midrule
        \multirowcell{3}{INT3} 
        & 0     & 29.51 & 24.96 & 17.10 & 15.27 & 13.15 \\
        & 10    & 29.59 & 24.98 & 17.06 & 15.29 & 13.15 \\
        & 100   & 29.58 & 25.00 & 17.09 & 15.37 & 13.15 \\
        \midrule
        \multirowcell{3}{INT2} 
        & 0     & 56.88 & 46.36 & 23.54 & 22.53 & 17.28 \\
        & 10    & 56.23 & 44.02 & 23.91 & 22.56 & 16.91 \\
        & 100   & 56.78 & 45.21 & 24.41 & 22.42 & 17.30 \\
        \bottomrule
        \end{tabular}
    \end{subtable}

\end{table*}

\begin{table*}[htb!]
    \renewcommand{\arraystretch}{1.0}
    \fontsize{7.8pt}{9.36pt}\selectfont
    \centering
    \caption{Quantization performance statistics (PPL $\downarrow$) of \aespa~on OPT models.}
    \vspace{-.2cm}
    \begin{tabular}{c c c c c c c c}
    \toprule
    Precision & Dataset & 125M & 350M & 1.3B & 2.7B & 6.7B \\
    \toprule
    \multirowcell{3}{INT4} 
    & Wiki-2 & 28.74 $\pm$ 0.139 & 22.65 $\pm$ 0.172 & 14.89 $\pm$ 0.066 & 12.34 $\pm$ 0.023 & 10.89 $\pm$ 0.058 \\
    & PTB    & 40.41 $\pm$ 0.264 & 32.36 $\pm$ 0.217 & 20.63 $\pm$ 0.128 & 18.24 $\pm$ 0.057 & 15.87 $\pm$ 0.034 \\
    & C4     & 27.25 $\pm$ 0.036 & 23.14 $\pm$ 0.014 & 16.33 $\pm$ 0.016 & 14.55 $\pm$ 0.005 & 12.81 $\pm$ 0.002 \\
    \midrule
    \multirowcell{3}{INT3} 
    & Wiki-2 & 33.01 $\pm$ 0.340 & 24.49 $\pm$ 0.068 & 15.87 $\pm$ 0.209 & 13.21 $\pm$ 0.064 & 11.19 $\pm$ 0.068 \\
    & PTB    & 46.26 $\pm$ 1.287 & 35.98 $\pm$ 0.321 & 22.17 $\pm$ 0.159 & 19.54 $\pm$ 0.109 & 16.33 $\pm$ 0.058 \\
    & C4     & 29.56 $\pm$ 0.043 & 24.98 $\pm$ 0.024 & 17.08 $\pm$ 0.021 & 15.31 $\pm$ 0.050 & 13.15 $\pm$ 0.004 \\
    \midrule
    \multirowcell{3}{INT2} 
    & Wiki-2 & 74.20 $\pm$ 2.931 & 52.95 $\pm$ 2.029 & 24.86 $\pm$ 0.930 & 22.32 $\pm$ 0.088 & 15.48 $\pm$ 0.363 \\
    & PTB    & 105.0 $\pm$ 5.495 & 77.64 $\pm$ 3.195 & 36.82 $\pm$ 0.766 & 32.24 $\pm$ 0.335 & 22.06 $\pm$ 0.667 \\
    & C4     & 56.63 $\pm$ 0.350 & 45.20 $\pm$ 1.171 & 23.95 $\pm$ 0.438 & 22.50 $\pm$ 0.076 & 17.17 $\pm$ 0.219 \\
    \bottomrule
    \end{tabular}

    \label{tab:results_seed_stats}
    
\end{table*}

\end{document}